\documentclass{article}

\usepackage{microtype}
\usepackage{graphicx}
\usepackage{subfigure}
\usepackage{booktabs} 

\usepackage{hyperref}

\usepackage[algo2e]{algorithm2e} 

\usepackage[accepted]{icml2024}

\usepackage{amsmath}
\usepackage{amssymb}
\usepackage{mathtools}
\usepackage{amsthm}
\usepackage{multirow}
\usepackage{comment}
\usepackage{color, colortbl}
\usepackage{enumitem}
\usepackage{scalerel}
\usepackage{lipsum}
\usepackage{duckuments}
\usepackage{adjustbox}
\usepackage{tcolorbox}

\usepackage[capitalize,noabbrev]{cleveref}
\usepackage[textsize=tiny]{todonotes}

\theoremstyle{plain}

\theoremstyle{definition}

\theoremstyle{remark}

\newcolumntype{P}[1]{>{\centering\arraybackslash}p{#1}}
\colorlet{shadecolor}{gray!30}
\definecolor{shamrockgreen}{rgb}{0.0, 0.62, 0.38}
\definecolor{urobilin}{rgb}{0.88, 0.68, 0.13}
\definecolor{sinopia}{rgb}{0.8, 0.25, 0.04}

\usepackage{xcolor,pifont}
\newcommand*\colourcheck[1]{%
  \expandafter\newcommand\csname #1check\endcsname{\textcolor{#1}{\ding{52}}}%
}
\colourcheck{shamrockgreen}
\colourcheck{urobilin}

\definecolor{commentcolor}{RGB}{110,154,155}   
\definecolor{defcolor}{RGB}{225,81,145}

\newcommand{\PyComment}[1]{\fontsize{10}{12}  \ttfamily \textcolor{commentcolor}{\# \textls[-100]{#1}}}  
\newcommand{\PyCode}[1]{\fontsize{10}{12}  \ttfamily \textcolor{black}{\textls[-100]{#1}}} 
\newcommand{\PyDef}[1]{\fontsize{10}{12}  \ttfamily \textcolor{defcolor}{\textls[-100]{#1}}} 

\icmltitlerunning{Slow and Steady Wins the Race: Maintaining Plasticity with Hare and Tortoise Networks}

\begin{document}

\twocolumn[
\icmltitle{Slow and Steady Wins the Race \\ Maintaining Plasticity with Hare and Tortoise Networks}

\begin{icmlauthorlist}
\icmlauthor{Hojoon Lee}{sch1}
\icmlauthor{Hyeonseo Cho}{sch2}
\icmlauthor{Hyunseung Kim}{sch1}
\icmlauthor{Donghu Kim}{sch1}
\icmlauthor{Dugki Min}{sch2}
\icmlauthor{Jaegul Choo}{sch1}
\icmlauthor{Clare Lyle}{comp1}
\end{icmlauthorlist}

\icmlaffiliation{sch1}{KAIST}
\icmlaffiliation{sch2}{Konkuk University}
\icmlaffiliation{comp1}{Deepmind}

\icmlcorrespondingauthor{Hojoon Lee}{joonleesky@kaist.ac.kr}

\icmlkeywords{Machine Learning, ICML}

\vskip 0.3in]



\printAffiliationsAndNotice{} 

\begin{abstract}
This study investigates the loss of generalization ability in neural networks, revisiting warm-starting experiments from \citet{ash2020warm}. Our empirical analysis reveals that common methods designed to enhance plasticity by maintaining trainability provide limited benefits to generalization. While reinitializing the network can be effective, it also risks losing valuable prior knowledge. 
To this end, we introduce the Hare \& Tortoise, inspired by the brain's complementary learning system. Hare \& Tortoise consists of two components: the Hare network, which rapidly adapts to new information analogously to the hippocampus, and the Tortoise network, which gradually integrates knowledge akin to the neocortex. 
By periodically reinitializing the Hare network to the Tortoise's weights, our method preserves plasticity while retaining general knowledge.
Hare \& Tortoise can effectively maintain the network's ability to generalize, which improves advanced reinforcement learning algorithms on the Atari-100k benchmark. The code is available at \url{https://github.com/dojeon-ai/hare-tortoise}.
\end{abstract}


\section{Introduction}
\label{introduction}

\begin{figure}[t]
\begin{center}
\includegraphics[width=0.95\linewidth]{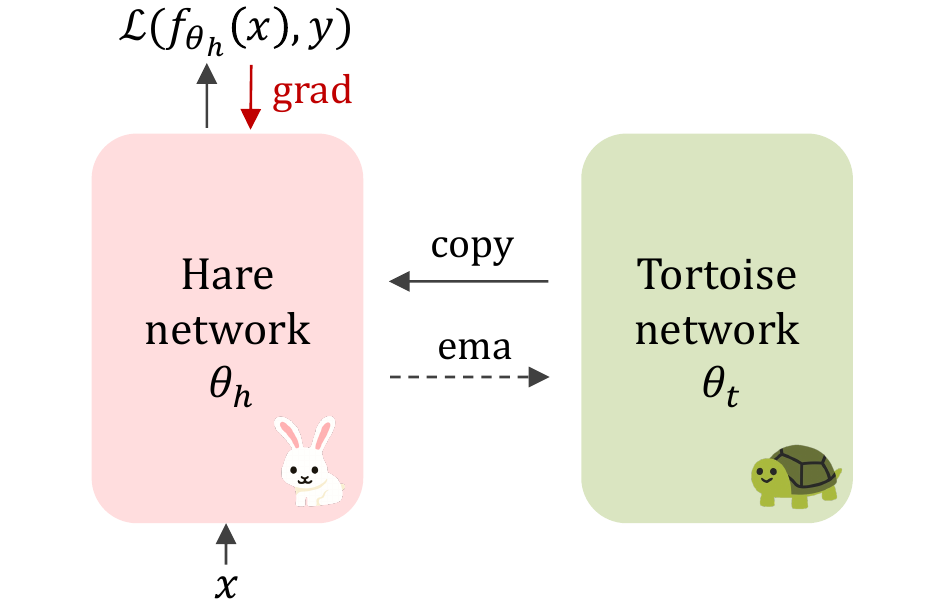}
\end{center}
\vspace{-2mm}
\caption{\textbf{Hare \& Tortoise architecture}. The Hare Network rapidly updates its weights for new data, while the Tortoise Network slowly integrates knowledge through an exponential moving average (ema) of the Hare's weights. Periodic reinitialization of the Hare Network to the Tortoise Network's weights ensures a balance between fast, fleeting adaptation and slow, steady generalization.}
\label{figure:overview_HnT}
\end{figure}
\vspace{-1mm}


In neural networks, \textit{plasticity} refers to the ability to learn and adapt to new data. 
Maintaining plasticity is crucial in continual learning and reinforcement learning, where networks must consistently adapt to incoming data. 
With the emergence of large-scale models like GPT-4 \cite{achiam2023gpt4} and Gemini \cite{team2023gemini}, understanding plasticity is becoming crucial, considering the substantial resources required to retrain these models from scratch.

Recent studies have shown that neural networks tend to lose plasticity as training progresses \cite{lyle2023understanding_plasticity, dohare2023maintaining}. 
In experiments where the dataset labels were periodically randomized, some networks gradually lost their ability to minimize their training loss.
This phenomenon has been linked to several factors such as the increasing distance from initial weights, emergence of gradient starvation, and accumulating dormant neurons \cite{lyle2023understanding_plasticity, dohare2023maintaining, lewandowski2023curvature}.  
To mitigate this problem, various methods have been developed \cite{lyle2022capacity, kumar2023regen, abbas2023plasticity_crelu, sokar2023dormant}, proving their efficacy in maintaining trainability throughout repetitive learning tasks.

Ultimately, the goal of maintaining plasticity is not just to memorize the new training data but to enable the network to generalize to unseen data.
However, it is not well understood whether, and if so how, improved trainability translates to better generalization. The first contribution of this paper is to conduct a rigorous empirical study into the relationship between these dual faces of plasticity.
To do so, we expand the warm-starting experiments from \citet{ash2020warm} by first training some networks on noisy dataset subsets and then retraining them on complete, noise-free datasets.
Here, we observed a consistent decline in generalization when models were initially trained on smaller and noisier subsets. 

On top of this setup, we investigate whether methods designed to maintain trainability can mitigate this decline.
Surprisingly, contrary to the belief in previous studies \cite{abbas2023plasticity_crelu, sokar2023dormant, lee2023plastic, lyle2023understanding_plasticity}, we found that while these methods maintain the network's ability to reduce training loss, they do not improve generalization.
Instead, we observe that standard training techniques such as L2 regularization \cite{krogh1991simple_l2} and data augmentation \cite{takahashi2019data_aug} tend to have a much greater effect on generalization. However, they are not sufficient to completely address the loss of generalization ability. 
Weight Re-initialization methods such as Shrink \& Perturb \cite{ash2020warm} emerge as the most effective solution, enhancing plasticity by periodically resetting and scaling weights.

While effective, resetting the network (even if only partially) is costly. 
In large models, resetting resembles a form of less intensive retraining and demands considerable effort to restore prior performance. When data access is limited due to privacy \cite{abadi2016differntial, mcmahan2017federated} or storage constraints \cite{shin2017continual_generative, smith2023rehersal_free}, it risks losing previously learned valuable information.
In the end, we seek a system that allows the network to quickly adapt to new data and overwrite spurious correlations without erasing all of its prior knowledge. In our efforts towards this goal, we draw inspiration from the human brain. 

Humans maintain plasticity through a dynamic interaction within the brain's complementary learning systems \cite{mcclelland1995why_complementary}. 
The human brain possesses two distinct memory modules: the fast-learning hippocampus and the slow-learning neocortex \cite{kumaran2016complementary}. The hippocampus rapidly encodes and transfers new information to the neocortex, which is responsible for storing and retaining long-term knowledge and skills.
While the hippocampus periodically forgets its knowledge to preserve the brain's plasticity, the neocortex serves as a long-term module to preserve general knowledge \cite{frankland2013hippocampal_forget}. 

Inspired by the complementary learning theory, we introduce a novel network architecture, Hare \& Tortoise, reflecting the brain's fast and slow learning mechanisms. 
As illustrated in Figure \ref{figure:overview_HnT}, the Hare network, like the hippocampus, rapidly updates information and explores the parameter space. On the other hand, the Tortoise network, akin to the neocortex, slowly integrates knowledge from the Hare network by an exponential moving average. To maintain plasticity, the Hare network is periodically reinitialized to the Tortoise network's weights, which naturally retains the generalizable knowledge of the slow-learning Tortoise.

Hare \& Tortoise consistently yields impressive results across various experimental setups.
In warm-starting and continual learning experiments, Hare \& Tortoise effectively maintain generalization abilities without any sudden performance drop after resets, showing competitive performance compared to reinitialization methods.
Furthermore, in reinforcement learning, Hare \& Tortoise can be seamlessly integrated into modern algorithms that employ momentum target networks.
By just periodically reinitializing the online network to the target network, Hare \& Tortoise enhances the sample efficiency of both DrQ \cite{kostrikov2020drq} and BBF \cite{schwarzer2023bbf} in the Atari-100k benchmark.


\section{Related Work}
\label{section:related_work}

\subsection{Loss of Plasticity}

The design of neural network architectures and initialization schemes that both train and generalize well has produced a rich scientific literature \citep{ glorot2010understanding, he2015delving, he2016deep}, allowing supervised training to reliably scale to immense model and dataset sizes \citep{kaplan2020scaling}. In recent years, however, many works have identified the insufficiency of standard optimization algorithms in non-stationary domains, such as reinforcement learning \citep{lyle2023understanding_plasticity}, continual learning \citep{kumar2023continual, dohare2023maintaining}, and lifelong learning \citep{sodhani2020toward}. In particular, these works identify a phenomenon known as \textit{loss of plasticity}, whereby neural networks progressively lose their ability to learn and adapt to new data.

Loss of plasticity, as noted by \citet{berariu2021study}, can be decomposed into two distinct factors: a reduced ability of networks to minimize the training loss on new data, \textit{trainability}, and a reduced ability to generalize well to unseen data, \textit{generalizability}. In the former case, \citet{lyle2022capacity} observed that neural networks can exhibit reduced performance over time on sequential memorization tasks, where no generalization is required. In the latter case, \citet{ash2020warm} observed reduced generalization performance after warm-starting on a subset of the training data, despite achieving zero training error.

Understanding the precise causes of plasticity loss for training and generalization remains unclear \cite{ma2023revisiting, lyle2023understanding_plasticity, lyle2024disentangling}, but efforts to mitigate it have been diverse. 
To enhance trainability, methods include maintaining active units \cite{abbas2023plasticity_crelu, elsayed2024addressing}, preventing gradient starvation \cite{gogianu2021spectral, lyle2022capacity, dohare2023maintaining}, using small batch size \cite{obando2024small}, and limiting deviation from initial weights \cite{lewandowski2023curvature, kumar2023regen}. 
For generalization, periodicaly reinitializing the network has been effective \cite{ash2020warm, zhou2022fortuitous, zaidi2023whendoesreinit, noukhovitch2023elastic, frati2023reset_continual}, particularly in data-efficient RL \cite{nikishin2022primacy, lee2023plastic, xu2023drm, nauman2024overestimation}.

Despite these efforts, the relationship between enhancing trainability and achieving better generalization under non-stationary learning conditions is not fully understood. Our research aims to bridge this gap by investigating whether modern neural networks suffer from a loss of trainability and whether enhancing trainability can counteract this decline.

\subsection{Complementary Learning System}

The Complementary Learning Systems (CLS) hypothesis conjectures that learning in the brain is mediated by two distinct systems: a fast learning system located in the hippocampus which rapidly incorporates new information, and a slower system in the neocortex which integrates this information into long-term knowledge
\cite{mcclelland1995why_complementary, kumaran2016complementary}.
Another key process in the brain is forgetting, which discards old information to make room for new knowledge, a critical component in maintaining neural plasticity \cite{frankland2013hippocampal_forget, gravitz2019forgotten, ryan2022forget_plasticity}.

In deep learning, the CLS principle has been applied to problem settings that require continual adaptation.
In continual learning, dual-network architectures have been designed to replicate the mechanisms of the hippocampus and neocortex. The hippocampal network, fast learner, undergoes standard supervised learning, while the neocortical network, slow learner, is trained by self-supervised objectives \cite{pham2021dualnet}, ensembles \cite{arani2022fast_slow_continual, pham2022fast_slow_timeseries}, or knowledge distillation \cite{gomez2024plasticity_distill}. Similarly, in RL, distinct architectural designs \cite{duan2016rl2, pritzel2017neuralepisodic} or decomposed value functions \cite{anand2023prediction_control} have been employed to echo this concept.

This work distinguishes itself from existing methods by explicitly integrating a forgetting mechanism to enhance the network's plasticity. Unlike previous studies based on the CLS principle, we emphasize the crucial role of forgetting to acquire new knowledge. In our approach, the Hare network is periodically reinitialized to the Tortoise network's weights, forgetting obsolete information while preserving generalizable knowledge within the Tortoise network.







\begin{figure*}[t]
\begin{center}
\includegraphics[width=0.94\linewidth]{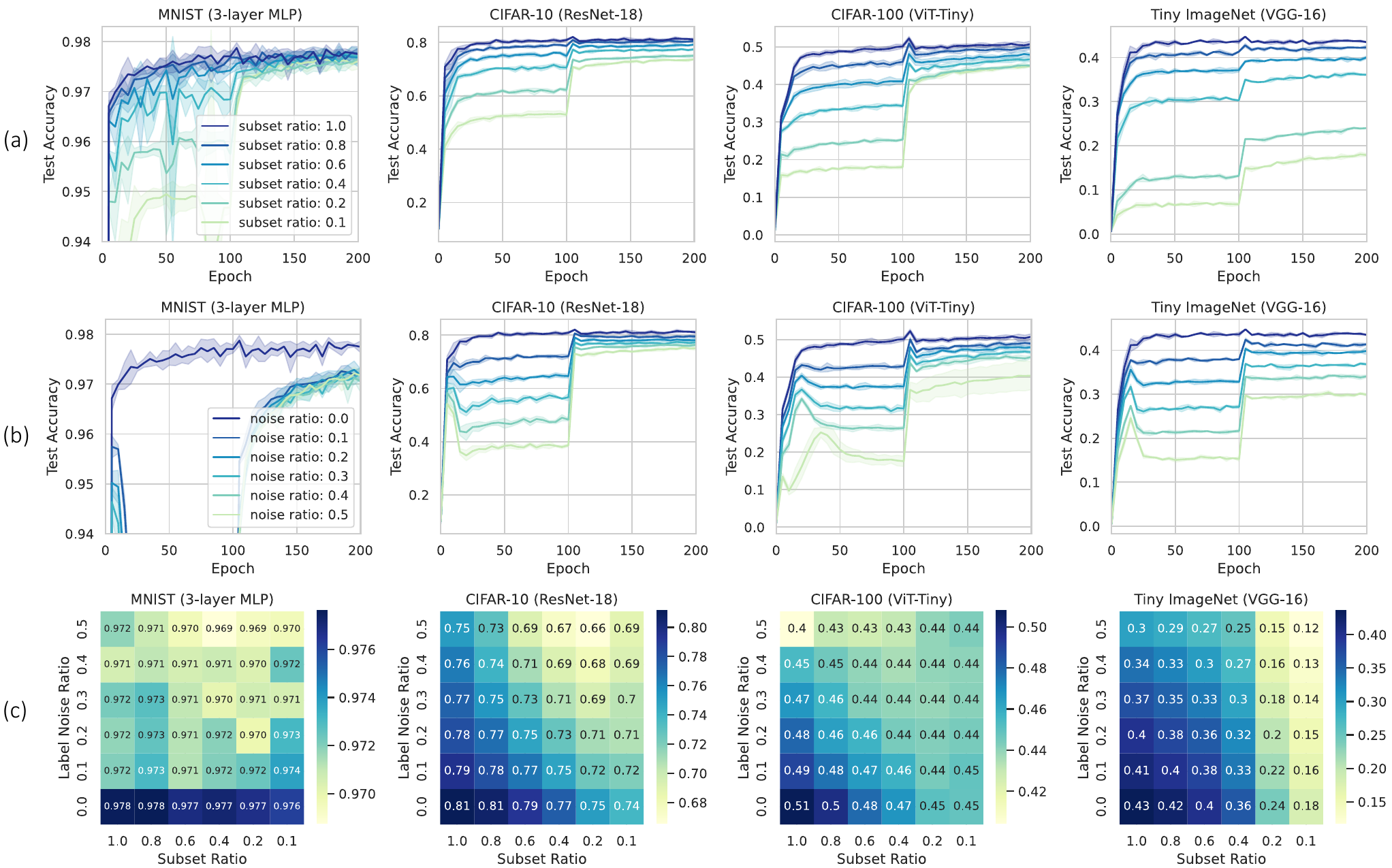}
\end{center}
\vspace{-2.1mm}
\caption{\textbf{Impact of Warm-Starting on Generalization.} 
\textbf{(a)} Shows a negative correlation between test accuracy and subset ratio without label noise. \textbf{(b)} Presents a negative correlation between test accuracy and label noise ratio, with a full dataset. \textbf{(c)} Presents the combined impact, indicating both reduced data size and increased label noise detrimentally affect generalization.}
\vspace{-1mm}
\label{figure:fig2}
\end{figure*}



\section{Investigating the Effect of Warm-Starting on Neural Network Generalization} \label{section3:warmstart}

Many methods in the literature that aim to mitigate the loss of plasticity focus on ensuring the network remains \textit{trainable}, meaning the network can continually minimize the training loss.
It is not guaranteed, however, that trainable networks necessarily avoid overfitting \citep{zhang2021understanding, xiao2020disentangling}. This section will use the warm-starting regime of \citet{ash2020warm} to disentangle the dual aspects of plasticity into \textit{trainability} and \textit{generalizability}. Our analysis will aim both to identify features of a task that increase the risk of overfitting and to assess whether existing approaches can maintain a network's \textit{generalizability}.

\subsection{Experimental Setup}
Adopting the experimental setup from \citet{ash2020warm, zaidi2023whendoesreinit}, we begin by training networks on noisy data subsets, followed by retraining it on complete, noise-free datasets.
This setup is particularly relevant to deep reinforcement learning using a temporal difference loss \cite{tesauro1995temporaldifference, sutton2018reinforcement}. Here, the network begins training with a small dataset containing noisy targets, followed by a gradual increase in data size and target accuracy.
Our experiments are designed to yield insight into neural networks' generalization abilities under analogous non-stationary learning dynamics.

\textbf{Dataset and Architecture.}
For our study, we select datasets and architectures that are widely used. 
We used the MNIST \citep{lecun1998mnist}, CIFAR-10, CIFAR-100 \citep{krizhevsky2009learning} and  Tiny ImageNet \citep{le2015tiny} datasets. 
For each dataset, we paired a network architecture of 3-layer MLP, ResNet-18 \citep{he2016deep}, ViT-Tiny \citep{dosovitskiy2020image}, and VGG-16 \citep{simonyan2014very}. 


\textbf{Warm-starting.}
Our networks were initially warm-started using a subset of the dataset, with varying ratios from $\{0.1, 0.2, 0.4, 0.6, 0.8, 1.0\}$, where 1.0 represents the entire dataset. Following \citet{zhang2021understanding}, we inject label noise by substituting true labels with random ones at varying ratios of $\{0.0, 0.1, 0.2, 0.3, 0.4, 0.5\}$. Subsequently, the network was trained on the complete dataset without label noise.

\textbf{Training Details.}
In both warm-starting and subsequent training phases, we employed the AdamW optimizer \citep{loshchilov2017decoupled} with L2 weight decay and a batch size of 256.
For each dataset, the learning rate was tuned via grid search, ranging from $\{0.01, 0.001, 0.0001, 0.00001\}$.  
Since the optimizer statistics within the Adam optimizer ($\beta_1, \beta_2$) can bias the optimization trajectory, we reinitialized the optimizer before proceeding to the subsequent training phase \cite{gogianu2021spectral, lyle2023understanding_plasticity, asadi2023resetting}. 

Training epochs were fixed to 100 for both phases, with a proportional increase in epochs during the warm-starting phase with subsets, to equalize the number of total gradient updates (e.g., 1000 epochs for 10\% of the dataset). 5 random seeds were used for experiments, except for Tiny ImageNet where 3 seeds were employed. 

\subsection{Effects of Warm-Starting on Generalizability}
\label{subsection:size_noise_effects}

First, we study the effects of warm-starting on generalization, w.r.t reduced data diversity and increased label noise.

In Figure \ref{figure:fig2}.(a), we examine the relationship between initial dataset size and test accuracy, without label noise. Our analysis reveals that reductions in dataset size impair generalization, as evidenced by lower test accuracy scores.
Figure \ref{figure:fig2}.(b) explores the impact of injecting varying levels of label noise with a complete dataset. We observe a significant degradation in generalization when these models are subsequently trained on noise-free datasets.
Figure \ref{figure:fig2}.(c) explores the combined effects of smaller subsets and increased label noise, showing that these factors exacerbate performance deterioration. 
Despite this decline, warm-started models achieved near-perfect training accuracy, comparable to models that were freshly initialized (Appendix \ref{appendix:additional_warm_start}).

Our analysis also reveals that both dataset type and network architecture influence the network's plasticity. For instance, warm-starting the VGG architecture on Tiny ImageNet resulted in a substantial loss compared to warm-starting an MLP architecture on the MNIST dataset. While both factors are important, the dataset had a more pronounced effect. This is further corroborated in Appendix \ref{appendix:dataset_architecture_impact}, where we conduct additional experiments with CIFAR-10 using VGG16 and Tiny ImageNet using ResNet-18.

\subsection{Effects of Optimizer on Generalizability}

\begin{figure}[H]
\vspace{-2mm}
\begin{center}
\includegraphics[width=1.0\linewidth]{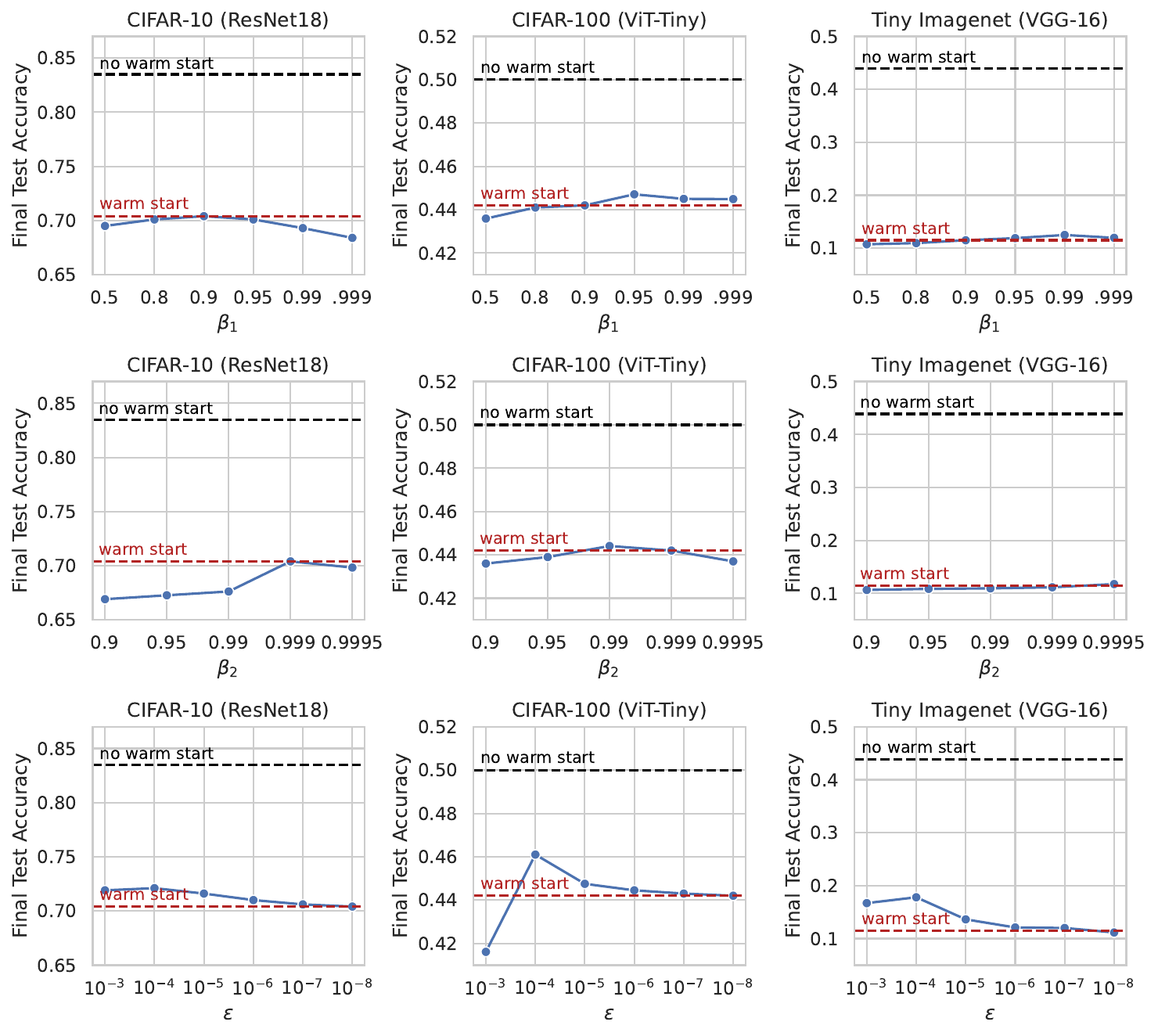}
\end{center}
\vspace{-3mm}
     \caption{\textbf{Effect of Optimizer Parameters.} We observed marginal improvements with varying $\beta_1$ and $\beta_2$. Larger $\epsilon$ alleviate generalization loss but are insufficient to address it entirely.}
\label{figure:adam}
\end{figure}

Modern neural networks commonly utilize adaptive optimizers like Adam \cite{kingma2014adam} or LARS \cite{you2019lars}, which adjust convergence speed through parameters such as $\beta_1$, $\beta_2$, and $\epsilon$. To assess the potential of these parameters in mitigating generalization loss during warm-starting, we conducted a grid search over $\beta_1$, $\beta_2$, and $\epsilon$ in Adam optimizer. Results are depicted in Figure \ref{figure:adam}.

\begin{figure*}
\begin{center}
\includegraphics[width=0.97\linewidth]{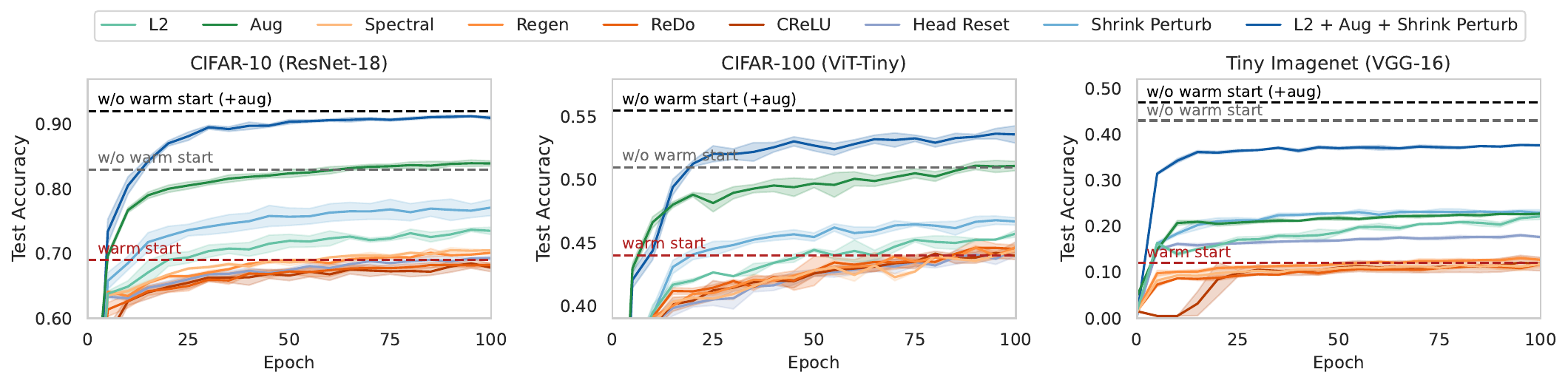}
\end{center}
\vspace{-3mm}
\caption{\textbf{Comparison of Existing Methods.} 
This figure presents a comparative analysis of test accuracies for different methods applied to networks warm-started with a 10\% subset ratio and 50\% label noise. Dashed lines indicate the performance of a warm-started network (lower bound) and a fresh network without warm-starting (upper bound). Generalizability methods (L2, Aug) are marked in green, Trainability methods (Spectral, Regen, ReDo, CReLU) in red, and Re-initialization methods (Head Reset, Shrink \& Perturb) in blue.}
\label{figure:baseline}
\end{figure*}

Our findings indicate that tuning $\beta_1$ and $\beta_2$ showed marginal improvements, with default values generally performing best ($\beta_1, \beta_2$ = (0.9, 0.999)). Decreasing momentum in Adam may slow convergence but does not effectively counteract generalization loss.

Conversely, larger epsilon values seemed to alleviate generalization loss, aligning with prior studies \cite{gogianu2021spectral, lyle2023understanding_plasticity}. This indicates that larger epsilon induces smoother gradients, mitigating the generalization aspect of plasticity loss. Notably, large epsilon values are commonly used in RL algorithms as implementation details \cite{hessel2018rainbow}, possibly contributing to this loss of generalizability. However, despite their benefits, they were insufficient to entirely address generalization loss.

\subsection{Enhancing Trainability is Insufficient for Maintaining Generalizability}
\label{sub_section:generalization_effectiveness}


Having observed a significant degradation in networks' generalization ability from warm-starting, we now investigate whether existing methods can mitigate this decline.
We categorized existing methods into three groups: Generalizability, Trainability, and Re-initialization. Generalizability includes approaches that aim to prevent overfitting and enhance generalization. Trainability involves methods designed to consistently minimize the training loss, which has been established to address the plasticity loss on the training side. Re-initialization encompasses methods that reinitialize parts of the network to their initial weight distribution.

We conduct these evaluations under the most challenging conditions observed in our earlier analysis: a subset ratio of 0.1 and a label noise ratio of 0.5. 
We include a vanilla warm-started network as a lower bound and a fresh network without warm-starting as an upper bound.

\textbf{Generalizability.} 
We investigated widely used techniques, such as L2 regularization \cite{krogh1991simple_l2} and Data Augmentation \cite{takahashi2019data_aug}. As illustrated in Figure \ref{figure:baseline}, L2 regularization demonstrated improved generalization across datasets, leading to enhanced test accuracy. Data Augmentation proved its effectiveness, yielding results similar to those of models without warm-starting. However, a performance gap persists between warm-started models with Data Augmentation and their non-warm-started counterparts, indicating persistent generalization loss.

\textbf{Trainability.} We explored various trainability methods, including Spectral decoupling \cite{pezeshki2021gradient}, Regenerative regularization \citep{kumar2023regen}, ReDo \citep{sokar2023dormant}, and CReLU \citep{abbas2023plasticity_crelu}.
While these methods accelerated the convergence of training loss, none of them enhanced generalization on unseen data.

We also examined the relationship between generalization performance and potential indicators of plasticity loss, such as weight magnitude, active unit fraction, and feature rank. However, none of these metrics show a consistent correlation with test accuracy, thus not elucidating the loss of generalization ability. For details, refer to Appendix \ref{appendix:metrics}.

\textbf{Re-initialization.}
We explore two methods: Head Reset \citep{nikishin2022primacy} and Shrink \& Perturb \citep{ash2020warm}.
For Head Reset, the MLP network after the encoder is reinitialized to its initial weight. While Head Reset has shown its effectiveness in RL literature \citep{nikishin2022primacy}, it only improves generalization over the other baselines in the VGG architecture, where the head network contains a relatively larger fraction of the total parameters (18\%). This observation aligns with RL setups, where nearly 90\% of the parameters are included in the Head network \cite{d2022sample_breaking}. Therefore, we conjecture that Head Reset may not scale effectively with architectures featuring larger and deeper encoders.

Shrink \& Perturb involve shrinking the network's weights towards their initial values and perturbing with Gaussian noise. For simplicity, we focus solely on the shrinking process with a shrink ratio set to 0.8, where 80\% of the weights come from the initial values and 20\% from the current weights.
For all experiments, Shrink \& Perturb significantly reduce plasticity loss and enhance generalization across all datasets. Further incorporation of standard generalization techniques (L2 + Aug + Shrink \& Perturb) narrows the gap between this approach and training a fresh network without warm-starting.
See Appendix \ref{appendix:baseline_details} for details on each method.
 
In summary, our findings reveal that while loss of trainability was often observed in small neural networks with prolonged training \cite{lyle2022capacity, abbas2023plasticity_crelu, kumar2023regen}, modern neural networks trained in the warm-starting regime generally do not suffer from this issue. Subsequently, enhancing trainability does not necessarily improve generalization in these cases. While certain generalization strategies (i.e., L2 regularization, data augmentation) showed moderate success, integrating them with weight reinitialization methods, particularly Shrink \& Perturb, proved most effective. 

However, naive reinitialization has certain limitations. While reinitialization recovers network plasticity, it concurrently loses valuable information. This drawback is particularly prominent when data access is limited due to privacy constraints or during the training process of larger models, where it exacerbates computational costs. 

\begin{tcolorbox}[boxsep=0pt,
                  left=8pt,
                  right=9pt]
\textbf{Takeaways:}
\begin{itemize}[leftmargin=10pt, topsep=4pt]
    \item Plasticity can be decoupled into \textit{trainability} (ability to train) and \textit{generalizability} (ability to generalize).
    \item Modern networks tend to retain \textit{trainability} but lose \textit{generalizability} after warm starting. 
    \item Warm-starting on smaller datasets with larger label noise exacerbates the loss of \textit{generalizability}.
    \item Varying optimizer parameters ($\beta_1, \beta_2, \epsilon$ in Adam) are insufficient to mitigate loss of \textit{generalizability}.
    \item Enhancing \textit{trainability} does not necessarily improve \textit{generalizability} when \textit{trainability} is retained.
    \item Standard regularization techniques, such as data augmentation and weight regularization, are partially effective in maintaining \textit{generalizability}.
    \item Reinitializing the network can recover \textit{generalizability}, though it risks losing learned information.
\end{itemize}
\end{tcolorbox}

\begin{figure*}
\begin{center}
\includegraphics[width=0.95\linewidth]{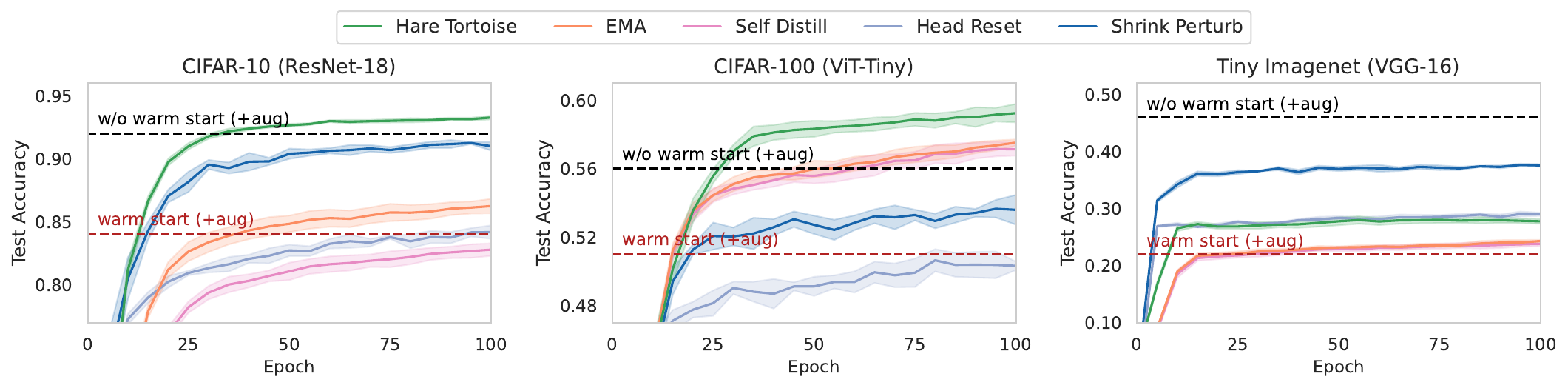}
\end{center}
\vspace{-4mm}
\caption{\textbf{Warm-Starting Results.} This graph presents the effectiveness of Hare \& Tortoise in warm-starting experiments, compared to EMA,  Self-Distillation, and Re-initialization methods. Hare \& Tortoise shows superior performance in CIFAR-10 (ResNet-18) and CIFAR-100 (ViT-Tiny), while reinitialization shows greater effectiveness in Tiny ImageNet (VGG-16) with severe generalization loss.}
\label{figure:fig4}
\end{figure*}

\section{Method}
\subsection{Hare \& Tortoise Architecture}

To maintain network plasticity while retaining valuable information, we present the Hare \& Tortoise architecture, inspired by the complementary learning systems in the human brain \cite{mcclelland1995why_complementary}. 
This architecture comprises two networks: the Hare network for rapid adaptation and the Tortoise network for stable knowledge consolidation.

\textbf{Hare network} ($h$): Inspired by the hippocampus, the Hare network rapidly adapts to new data by adjusting its parameters based on input-output pairs at each training step.

\textbf{Tortoise network} ($t$): Imitating the neocortex, the Tortoise network consistently accumulates knowledge over time by momentum updates \cite{tarvainen2017mean} from the Hare network. This ensures slow and steady integration of information acquired from the Hare network.

\subsection{Training Process}

The training process of the Hare \& Tortoise involves distinct yet interconnected updates to the networks.

At each training step, the Hare network's parameters, $\theta_h$, are updated by a gradient-based optimization algorithm. Given an input $x$ and its output $y$, the network is updated as:
$$ \theta_h \leftarrow \theta_h - \alpha \nabla_{\theta_h} \mathcal{L}(h(x; \theta_h), y) $$
where $\alpha$ denotes the learning rate, and $\mathcal{L}$ is the loss function.

Subsequently, the parameters of the Tortoise network, $\theta_t$, are updated using an exponential moving average based on the Hare network's parameters:
$$ \theta_t \leftarrow \mu \theta_t + (1 - \mu) \theta_h$$
where $\mu$ denotes the momentum.

To maintain plasticity in the Hare network, we incorporate a soft form of Re-initialization by periodically resetting its parameters to the Tortoise network's parameters:
$$ \theta_h \leftarrow \theta_t$$
This updating scheme has a number of desirable properties. The update rate for the Tortoise network $\mu$ gives a means of controlling how far from the initialization the parameters can deviate. By setting a small update rate (i.e., large momentum value), the Tortoise network can gradually incorporate useful information from the Hare network. 
Since the Hare network re-starts from the Tortoise weights upon every reset, it can converge more quickly than if it had been randomly initialized. Hard resets of the Hare network also provide an opportunity to escape from bad local minima which may contain spurious features. 

We provide pseudocode for the method in Algorithm \ref{algo:Hare and Tortoise}.

\subsection{Implementation}

In all experimental setups,  we utilized the Tortoise network for predictions, yielding stable and generalized outputs by ensembling the Hare network's parameters over time \cite{tarvainen2017mean, anonymous2023exponential}.

\label{pseudocode}
\begin{center}
\begin{minipage}{1.0\linewidth}
\vspace{-3mm}
\begin{algorithm}[H]
    \spaceskip=3pt plus 0pt minus 1pt
    \caption{Hare \& Tortoise Pseudocode (Pytorch-like)}
        \PyComment{h: hare network} \\
        \PyComment{t: tortoise network} \\
        \PyComment{m: momentum} \\
        \PyComment{r: reset interval} \\
        \PyDef{for} \PyCode{step, (x, y)} \PyDef{in} \PyCode{enumerate(loader):}    
        
        \Indp
        \PyComment{update hare network} \\
        \PyCode{logits = h(x)} \\
        \PyCode{loss = loss\_fn(logits, y)} \\
        \PyCode{loss.backward()} \\
        \PyCode{optimizer.step(h.params)} \\
        \PyComment{update tortoise network} \\
        t.parms = m*t.params + (1-m)*h.params \\
        \PyDef{if} \PyCode{step \% r == 0:}         

        \Indp
        \PyCode{h.params = t.params}
\label{algo:Hare and Tortoise}
\end{algorithm}
\end{minipage}
\end{center}

For our warm-starting and continual learning experiments, we fixed the momentum value of $\mu= 0.999$ and set the reset interval to occur every $10$ epoch across all experiments.

In reinforcement learning setup, the Hare \& Tortoise architecture can be seamlessly integrated into the modern algorithms that employ momentum target networks \cite{d2022sample_breaking, schwarzer2023bbf, hansen2023tdmpcv2, fujimoto2023td7}.
Here, the online network acts as the Hare network, and the target network serves as the Tortoise network. 
This integration requires only a single line of additional code, which periodically reinitializes the online network to the target network.
Following the previous setups \cite{d2022sample_breaking, schwarzer2023bbf}, we used a momentum value of $\mu=0.995$ and set the reset interval to $4000$ for every gradient update step.

\begin{figure*}
\begin{center}
\includegraphics[width=0.97\linewidth]{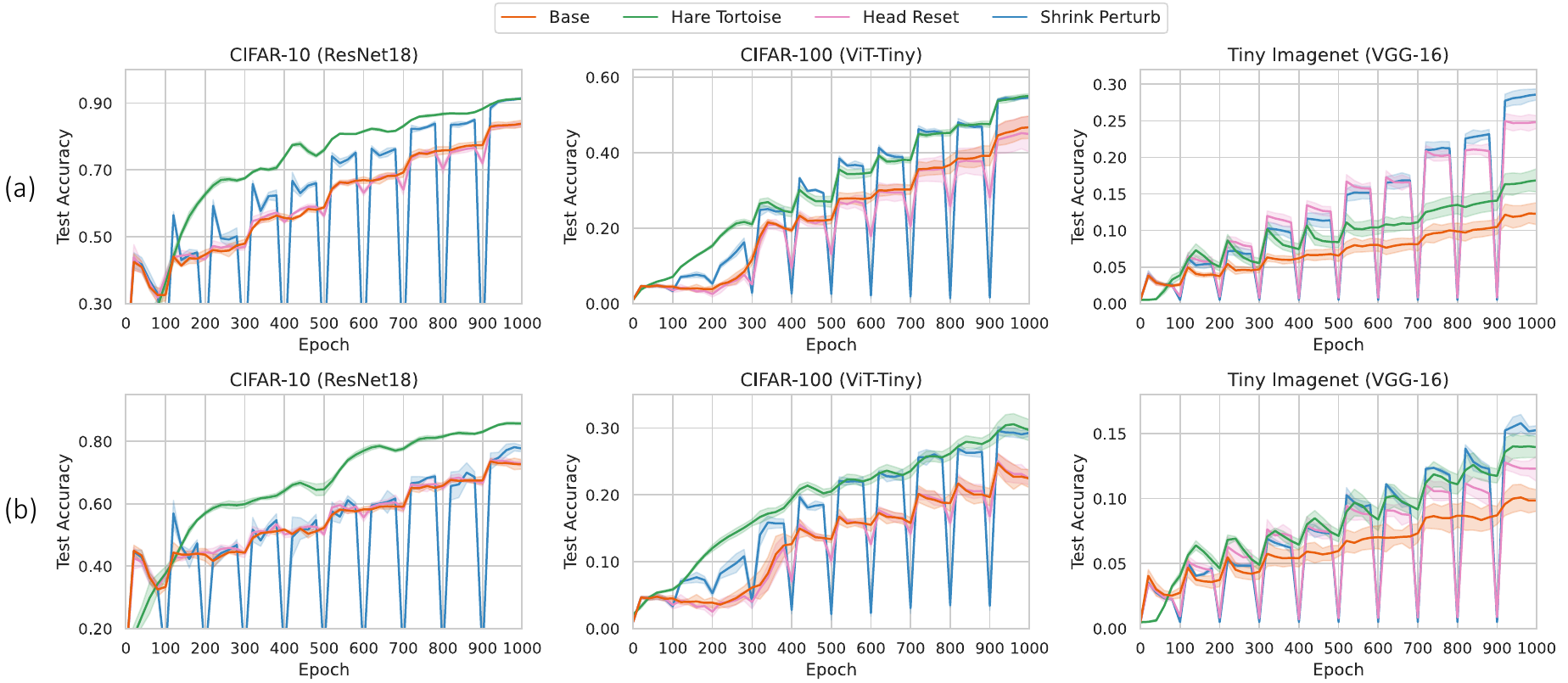}
\end{center}
\vspace{-1mm}
\caption{\textbf{Continual Learning Results.} The dataset was divided into 10 chunks, each undergoing 100 epochs of training with label noise decreasing from 0.5 to 0.0. \textbf{(a)} Shows outcomes with full dataset access, where Hare \& Tortoise consistently outperform Shrink \& Perturb in CIFAR-10 and CIFAR-100, but are less effective in Tiny ImageNet with significant generalization loss. \textbf{(b)} Illustrates results under limited access, with a buffer size of 5000. Here, Hare \& Tortoise shows strong performance in CIFAR-10 and CIFAR-100 and narrows the gap with Shrink \& Perturb in Tiny ImageNet, demonstrating its effective knowledge retention.}
\label{figure:continual}
\end{figure*}

\section{Experiments}


We evaluate the effectiveness of Hare \& Tortoise on maintaining generalizability in three distinct scenarios: warm-starting, continual learning, and reinforcement learning.

\subsection{Warm-Starting} 
\label{subsection:HnT_warmstart}

Following our warm-starting experiments in Section \ref{section3:warmstart}, we assessed the Hare \& Tortoise architecture against reinitialization methods like Shrink \& Perturb and Head Reset. We integrated L2 Regularization and Data Augmentation in all experiments due to their proven generalization benefits.

We also compared our method with Exponential Moving Average (EMA) and Self-Distillation \cite{tarvainen2017mean, allen2020understanding_self_distill}, to evaluate the impact of reinitialization of the Hare network on the Tortoise network. The Hare \& Tortoise architecture is a unique form of self-distillation, where the Tortoise guides the Hare to not deviate, but allows more freedom to explore the optimization landscape. For both EMA and Self-Distillation, we maintained a constant momentum value of $\mu = 0.999$, and tuned the distillation strengths from $\{1, 5, 10\}$.

As illustrated in Figure \ref{figure:fig4}, using EMA consistently brings benefits, particularly in CIFAR-100 using ViT-Tiny architecture.
This improved performance is likely due to ViT's convergence on sharp loss curvature, which further benefits from EMA's parameter smoothing \cite{chen2021when_vit, park2022how_vit}. 
The Hare \& Tortoise consistently outperformed both EMA and Self-Distillation, which verifies its effectiveness in maintaining plasticity and encouraging the Hare network's exploration in the optimization landscape.

Hare \& Tortoise not only outperformed other reinitialization strategies but also surpassed the performance of fresh networks in CIFAR-10 with ResNet-18 and CIFAR-100 with ViT-Tiny. However, its effectiveness was less pronounced on Tiny ImageNet with VGG-16. This implies that a radical form of reinitialization can be particularly beneficial in scenarios with substantial generalization loss.

To further understand the robustness of Hare \& Tortoise, we conducted an ablation study by varying momentum value and reset interval (Appendix \ref{appendix:ablation_studies}). 
The results indicate that Hare \& Tortoise is robust to hyperparameter choices, with larger momentum generally improving performance.

\subsection{Continual Learning}

Although Hare \& Tortoise architecture exhibited strong performance in our warm-starting experiments, the Shrink \& Perturb showed notable effectiveness in Tiny ImageNet. However, standard reinitialization methods face challenges in continual learning scenarios, which causes abrupt drops in online accuracy and increases computational costs for performance recovery. Moreover, when data access is limited, they risk losing valuable information as well.

To assess whether the Hare \& Tortoise can overcome these challenges, we extended our warm-starting experiments into a continual learning framework, which extends the training into 10 phases. Each phase had a fixed subset ratio of 0.1 and label noise was decreased from 0.5 to 0, using 100 epochs for training. Two data access scenarios were tested: one with full dataset access and another with limited access with a buffer size of 5000.

As depicted in Figure \ref{figure:continual}.(a), under full data access, Hare \& Tortoise excelled in CIFAR-10 and CIFAR-100, avoiding the abrupt performance drops associated with reinitialization. However, in Tiny ImageNet, where generalization loss is severe, reinitialization p roved to be a more effective choice, despite its periodic performance drops.

In the limited data access scenario (Figure \ref{figure
}.(b)), Hare \& Tortoise maintained strong performance on CIFAR-10 and CIFAR-100. The gap between Hare \& Tortoise and other reinitialization strategies was smaller on Tiny ImageNet, indicating that while naive reinitialization can be beneficial, it may also result in the loss of useful information for future tasks. Conversely, Hare \& Tortoise demonstrated better long-term information retention, particularly on CIFAR-10.

\begin{table*}[h]
\begin{center}
\caption{\textbf{Atari-100k Results.} BBF results without Hare \& Tortoise come from the original paper \cite{schwarzer2023bbf}. All the other experiments, including DrQ, were conducted based on their original code and averaged over 5 random seeds with a replay ratio of 2.}
\vspace{1.2ex}
\resizebox{0.95 \textwidth}{!}{
\begin{tabular}{l c c c c c c c c c c c c}
\toprule
Algorithm
& Architecture
& S\&P
& HR
& H\&T
& SSL
& GPU hours
& IQM    $\uparrow$ 
& Median $\uparrow$ 
& Mean   $\uparrow$ 
& OG     $\downarrow$ \\
\midrule \\[-2.5ex]

\multirow{6}{*}{DrQ \cite{kostrikov2020drq}}
& \multirow{6}{*}{3-layer ConvNet}
& -
& -
& -
& -
& \multirow{6}{*}{0.5}
& 0.243
& 0.193
& 0.468 
& 0.642  \\

&
& \checkmark 
& -
& -
& -
& 
& 0.139
& 0.138
& 0.458 
& 0.728  \\

& 
& -
& -
& \checkmark
& -
& 
& 0.287
& 0.260
& 0.471 
& 0.617  \\

& 
& -
& 20k 
& -
& -
& 
& \textbf{0.332}
& 0.254
& \textbf{0.694} 
& \textbf{0.580}  \\

& 
& -
& 40k
& -
& -
& 
& 0.288
& 0.241
& 0.532 
& 0.607  \\

& 
& -
& 40k
& \checkmark
& -
&
& \textbf{0.328}
& \textbf{0.329}
& 0.584 
& \textbf{0.583}  \\

\cmidrule{0-10} \\[-2.5ex]

\multirow{3}{*}{BBF \cite{schwarzer2023bbf}}
& \multirow{3}{*}{15-layer ResNet}
& \checkmark
& \checkmark
& -
& -
& \multirow{2}{*}{1.4}
& 0.826
& 0.711
& 1.737
& 0.397  \\

& 
& \checkmark
& \checkmark
& \checkmark
& -
& 
& 0.891
& \textbf{0.749}
& 1.719 
& \textbf{0.372} \\

& 
& \checkmark
& \checkmark
& -
& \checkmark
& 2.8
& \textbf{0.940}
& \textbf{0.755}
& \textbf{2.175} 
& 0.377  \\
\bottomrule 
\end{tabular}}
\label{table:sota_comparison}
\end{center}
\vspace{-1mm}
\end{table*}

\subsection{Reinforcement Learning}
Finally, we evaluate Hare \& Tortoise (H\&T) in reinforcement learning setup,  analyzing its performance with two algorithms on the Atari-100K benchmark \cite{kaiser2019atari100k}. The first, DrQ \cite{kostrikov2020drq}, follows \citet{mnih2015human}'s architecture, featuring three convolutional layers followed by two fully connected layers, and includes random cropping augmentation. The second, BBF \citep{schwarzer2023bbf}, uses a ResNet-style architecture \citep{espeholt2018impala} with a custom reset protocol, higher replay ratio, and its use of auxiliary self-supervised objective \cite{schwarzer2020spr}.

We integrated Hare \& Tortoise into these algorithms by reinitializing the online network to the target network every 4000 steps, which only requires a single line of additional code.
Since the hyperparameters of the BBF were carefully tuned based on reset intervals, we integrated Hare \& Tortoise with reinitializations but intentionally excluded the self-supervised objective. 
This decision was based on our interpretation that while BBF's self-supervised objective aids in understanding temporal dynamics, it simultaneously constrains the online network's divergence from the momentum target network, similar to Hare \& Tortoise. 
Thus, we aimed to isolate the effects of this constraint and assess the effectiveness of Hare \& Tortoise.


Table~\ref{table:sota_comparison} presents our experimental results. Adding Hare \& Tortoise in DrQ led to a modest improvement in the median score. However, the most effective approach was resetting the head every 20,000 steps, likely due to the limited depth of DrQ’s encoder and the head network's significant parameter portion (90\%).
While a 40,000-step interval Head Reset wasn’t wholly effective in mitigating plasticity loss, combining it with Hare \& Tortoise achieved comparable performance to the 20,000-step interval Head Reset.

For BBF, adding Hare \& Tortoise significantly improved both IQM and OG scores, without incurring any extra computational costs.
While Hare \& Tortoise's performance on IQM and Mean scores was lower than using self-supervised learning objective, it surpassed OG scores and required only half of the computational resources.
It's important to note that we did not adjust any original hyperparameters, including the momentum value. Therefore, 
with further optimization and exploration, we believe that Hare \& Tortoise has great potential to effectively alleviate plasticity loss.

\section{Conclusion and Future Work}

In the research community, it is widely believed that enhancing a neural network's trainability will naturally improve its generalizability in continuous learning scenarios \cite{lee2023plastic, sokar2023dormant, abbas2023plasticity_crelu, lyle2023understanding_plasticity, lyle2024disentangling,kumar2023regen, dohare2023maintaining}. Past studies have shown that neural networks trained continuously over long periods may struggle to minimize training loss, leading to reduced generalization. As a result, efforts have been made to improve generalizability by focusing on trainability. However, these studies often use smaller network architectures (e.g., 3 to 5 layers) compared to the deep and large networks used today.

Our study investigates whether maintaining trainability is sufficient to protect generalizability
with modern datasets and architectures. By revisiting warm-starting experiments \cite{ash2020warm} with various modern network architectures, we have not observed trainability issues even with prolonged training on small, noisy labeled datasets. Furthermore, methods aimed at enhancing trainability did not improve generalization, often leading to overfitting. Rather, simply reinitializing the network proved effective, despite the potential risk of losing valuable knowledge.

To address this, we developed the Hare \& Tortoise algorithm. It combines two types of networks: the Hare, which optimizes weights rapidly, and the Tortoise, which updates weights slowly by momentum averaging. The Tortoise’s slowly updated weights serve as starting points for periodic resets, which helps the Hare to rapidly adapt and escape from bad local minima.
It outperformed Shrink \& Perturb in continual learning experiments and enhanced the efficacy of state-of-the-art reinforcement learning algorithms.

However, there's still much room to explore. A key unanswered question is why warm-started models fail to generalize new tasks. The Hare \& Tortoise offers insights into this question; the Tortoise's effective performance implies that certain regions in the optimization landscape can generalize well on current tasks while preserving their generalizability on new tasks. Future improvements should focus on identifying and exploiting these regions, possibly reducing or removing the need for hard reinitializations.

\section*{Acknowledgements}

This work was supported by Institute for Information \& communications Technology Promotion (IITP) grant funded by the Korea government (MSIT) (No.RS-2019-II190075 Artificial Intelligence Graduate School Program (KAIST), the National Research Foundation of Korea (NRF) grant funded by the Korea government (MSIT) (No. NRF-2022R1A2B5B02001913 and No. 2021R1A2C209494311).

Also, we would like to express our gratitude to David Abel for his valuable feedback on this paper.

\section*{Impact Statement}

Our research focuses on enhancing neural network plasticity, which is important to develop intelligent robotics. This enhancement allows intelligent robots to effectively learn, adapt, and respond to dynamic environments, leading to improved efficiency in manufacturing, logistics, and autonomous driving.

\bibliographystyle{icml2024}
\bibliography{main}


\newpage

\appendix
\onecolumn

\section{Implementation Details for Warm-Starting}
\label{appendix:warmstart_details}

This section outlines the implementation details of our exploration into the impact of warm-starting on neural network performance, as discussed in Section \ref{section3:warmstart}. Our study encompasses a variety of datasets and neural network architectures.

\subsection{Datasets}
Our study utilized four datasets, each with distinct complexity and characteristics, to thoroughly evaluate the effects of warm-starting. The following datasets were used:

\textbf{MNIST} \cite{lecun1998mnist}: Contains 70,000 grayscale images (60,000 for training, 10,000 for testing), with each image being a 28x28 pixel representation of handwritten digits across 10 classes. MNIST is a basic dataset commonly used to benchmark machine learning algorithms due to its simplicity.

\textbf{CIFAR-10} \cite{krizhevsky2009learning}: Consists of 60,000 color images (50,000 for training, 10,000 for testing) across 10 classes, each image sized at 32x32 pixels. This dataset features objects and animals, offering a higher level of classification challenge than MNIST because of its color diversity and intricate patterns.

\textbf{CIFAR-100} \cite{krizhevsky2009learning}: Similar in structure to CIFAR-10 but with 100 classes, totaling the same number of 32x32 color images but divided among more categories, with each class having 600 images. CIFAR-100 raises the complexity by requiring finer differentiation between a broader array of classes.

\textbf{Tiny ImageNet:} A scaled-down version of the ImageNet dataset \cite{russakovsky2015imagenet}, Tiny ImageNet includes 100,000 training and 10,000 testing images, resized to 64x64 pixels, across 200 classes. It challenges models on a much wider range of classes, significantly more than the MNIST or CIFAR datasets.

\begin{table}[h]
\caption{Description of Dataset.}
\vspace{2mm}
\label{table:Datasets}
\begin{center}
\begin{tabular}{lcccc}
\toprule
Datasets & MNIST & CIFAR-10 & CIFAR-100 & Tiny ImageNet \\
\midrule
Type & Grayscale & Color & Color & Color\\
Image size & $28\times 28$ & $32\times 32$ & $32\times 32$ & $64\times64$\\
\# Classes & 10 & 10 & 100 & 200 \\
Train size & 70,000 & 50,000 & 50,000 & 10,0000  \\
Test size &  10,000 & 10,000 & 10,000 &  10,000 \\

\bottomrule
\end{tabular}
\end{center}
\end{table}

\subsection{Architectures}
In our research, we employed well-known neural network architectures, each paired with a specific benchmark dataset. These combinations include:

\textbf{3-layer MLP:} Used for analyzing the MNIST dataset, this architecture consisted of a 3-layer Multi-Layer Perceptron (MLP). Each hidden layer contained 100 units, and the model comprised a backbone (first and second layers) and a head (last fully connected layer). Hyperparameters for this model included a learning rate of 0.001, a weight decay of 0.0001, and batch normalization in the backbone.

\textbf{ResNet-18} \cite{he2016deep}: Employed for CIFAR-10, this architecture is known for its deep residual learning framework and residual connections. To optimize for computational efficiency and input resolution, we removed the stem layers and utilized specific hyperparameters, including a learning rate of 0.001, a weight decay of 0.00001, and batch normalization.

\textbf{ViT-Tiny} \cite{dosovitskiy2020image}: Utilized for CIFAR-100, this model was adapted from the original DeiT-Ti model \cite{touvron2021training}. It featured a 4x4 patch size, an embedding dimension of 192, three attention heads, and a depth of twelve layers. We used a learning rate of 0.003, weight decay of 0.05, layer normalization, and a dropout rate of 0.1.

\textbf{VGG-16} \cite{simonyan2014very}: Selected for Tiny ImageNet, this model is known for its traditional ConvNet architecture, comprising several consecutive convolutional layers followed by fully connected layers with hidden dimensions of 1024. Hyperparameters included a learning rate of 0.001, weight decay of 0.0001, batch normalization, and no dropout.

Learning rates and weight decay were meticulously determined through grid search.

\subsection{Training Details}

In this section, we provide a comprehensive overview of the hyperparameters and configurations for each baseline model. We begin by presenting the base hyperparameters, which are consistently applied across all models and experiments. These global hyperparameters are summarized in Table \ref{table:baseline_hyperparameter}.

\begin{table}[h]
\caption{Base Hyperparameters for Warm-Starting.}
\vspace{2mm}
\label{table:baseline_hyperparameter}
\begin{center}
\begin{tabular}{lr}
\toprule
Hyperparameter & Value \\
\midrule
Epochs & 100 \\
Optimizer & AdamW \\
Optimizer Hyperparameters ($\beta_1$, $\beta_2$) & (0.9, 0.999) \\
Batch Size & 256 \\
Learning Rate Scheduler & Warmup \\
Warmup Ratio & 0.1 \\
Initial Learning Rate &  0.0 \\
Gradient Clip. & 0.5\\
\bottomrule
\end{tabular}
\end{center}
\end{table}

Our warm-starting approach, following the methodology outlined by \citet{ash2020warm}, involves initializing the network with subsets of the original dataset. These subsets are created at ratios of 0.1, 0.2, 0.3, 0.4, 0.6, 0.8, and 1.0. To ensure an equivalent number of updates across different initializations, we adjust the training epochs for each subset, setting them at 1000, 500, 250, 166, 125, and 100, respectively. Furthermore, we incorporate label noise into our experiments as described by \citet{zhang2021understanding}. This entails replacing the original labels with uniformly distributed random labels.

To ensure the robustness and reliability of our results, each experimental setup is repeated with multiple random seeds. Specifically, we utilize 5 random seeds for all datasets and models, except for Tiny ImageNet, where we employ 3 random seeds for computational efficiency.

For all experiments, we used an NVIDIA RTX 3090 GPU for neural network training and a 32-core AMD EPYC 7502 CPU for multi-threaded tasks.

\newpage

\subsection{Baseline Methods}
\label{appendix:baseline_details}
In this section, we provide details of baseline methods and the hyperparameters we used in Section \ref{sub_section:generalization_effectiveness}.

\textbf{L2 Regularization:}  L2 Regularization is a widely adopted technique in neural network training. It works by adding a penalty term to the loss function, which is proportional to the square of the magnitude of the network weights. This penalty encourages the model to learn simpler patterns, thus increasing generalizability. In our experiments, we tested L2 regularization with varying intensities $\{1.0, 0.1, 0.01, 0.001, 0.0001\}$. 

\textbf{Data Augmentation:} We applied data augmentation techniques, including horizontal flipping and random cropping with 4-pixel padding, to the original datasets for improved model generalization \cite{laskin2020rad, kostrikov2020drq}.

\textbf{Concatenated Rectified Linear Units (CReLU)}: 
CReLU is an activation function introduced by \citet{shang2016understanding}, which combines the positive and negative parts of the ReLU function as $\left[\text{ReLU}(x), -\text{ReLU}(-x)\right]$. It has been demonstrated to be effective in preserving trainability in continual reinforcement learning \citet{abbas2023plasticity_crelu}. We incorporated CReLU in the head of each model, reducing the parameter count by half in each hidden layer. For the 3-Layer MLP, we added CReLU in every activation function.

\textbf{Recycling Dormant Neurons (ReDo):} \citet{sokar2023dormant} identified the Dormant Neuron Phenomenon, where certain neurons exhibit significantly reduced activity levels during training, negatively impacting trainability. ReDo periodically scans all layers for $\tau$-dormant neurons, reinitializes their weights, and sets their outgoing weights to zero. We tested various $\tau$ values including $\{0.01, 0.02, 0.05, 0.1, 0.2, 0.5\}$, with reinitialization applied every 5 epochs.

\textbf{Regenerative Regularization (Regen):} Developed by \citet{kumar2023regen}, this method embeds L2 regularization towards initial parameters in the loss function. It has shown effectiveness in retaining plasticity during continual learning by implicitly and smoothly resetting low-utility weights. We tested different regeneration rates $\{1.0, 0.1, 0.01, 0.001, 0.0001\}$, and used the best value for each experiment.

\textbf{Spectral Decoupling (Spectral):} 
Spectral Decoupling, as discussed by \citet{pezeshki2021gradient}, addresses the Gradient Starvation issue in over-parameterized neural networks. Gradient Starvation is a phenomenon in which the network disproportionately focuses on a few features, neglecting others which leads to sub-optimal utilization of predictive features and affects trainability. Spectral Decoupling adds an L2 penalty to the network's logits, promoting a more balanced feature learning. We optimized the spectral rate from various values $\{1.0, 0.1, 0.01, 0.001, 0.0001, 0.00001\}$.

\textbf{Shrink \& Perturb:} The study by \citet{ash2020warm} discovered that training neural networks with only half of the dataset initially can negatively affect their generalizability. To address this, they introduced the Shrink \& Perturb method. This approach involves periodically reducing the current weights (shrinking) and introducing small random noises (perturbation). In our experiment, we tested shrinkage values of $\{0.0, 0.2, 0.4, 0.6, 0.8, 1.0\}$ and used the optimal value to be 0.8. This process was used before proceeding to the subsequent training phase.

\textbf{Head Reset:} The Primacy Bias refers to an agent's tendency to overfit earlier experiences. This often results in a loss of learning ability in Reinforcement Learning, as described by \cite{nikishin2022primacy}. To address this issue, it has been discovered that regularly resetting the agent's head can be effective. In our experiments, we reinitialized the head network before proceeding to the subsequent training phase.



\newpage
\section{Experimental Results of Warm-Starting}
\label{appendix:additional_warm_start}
\subsection{Train and Test Accuracies}

In Section 3, we detail our warm-start experiments, which assess model performance across various subset ratios and label noise levels, with results summarized in Figure \ref{figure:fig2}.(c). The experiments in Table \ref{table:warmstart_train} yield near-perfect training accuracies for MNIST, CIFAR-10, and Tiny ImageNet, as shown in Table 1, indicating robust training irrespective of subset or noise variations. CIFAR-100 also demonstrates high training accuracy across on par with a freshly initialized network.

However, Table \ref{table:warmstart_test} highlights a decline in test accuracies compared to training, underscoring a gap between the models' ability to learn from training data and their generalization to unseen data. This contrast points to the challenges in maintaining model generalizability across different experimental setups.




\begin{table}[h]    
\caption{\textbf{Final Train Accuracy.} The overall train accuracy is similar regardless of the subset ratio and label noise ratio.} 
\vspace{2mm}
\label{table:warmstart_train}
\begin{center}
\resizebox{0.98 \textwidth}{!}{
\begin{tabular}{cccccc}
\toprule
Subset Ratio & Label Noise & MNIST (3-layer MLP) & CIFAR-10 (ResNet-18) & CIFAR-100 (ViT-Tiny) &Tiny ImageNet (VGG-16) \\
\midrule
0.1 & 0.0 & 1.00$\pm$0.0003 & 1.00$\pm$0.0006 & 0.91$\pm$0.0014 & 0.98$\pm$0.0011 \\
0.1 & 0.1 & 1.00$\pm$0.0002 & 1.00$\pm$0.0002 & 0.91$\pm$0.0019 & 0.98$\pm$0.0018 \\
0.1 & 0.2 & 1.00$\pm$0.0002 & 1.00$\pm$0.0004 & 0.91$\pm$0.0020 & 0.98$\pm$0.0002 \\
0.1 & 0.3 & 1.00$\pm$0.0003 & 1.00$\pm$0.0004 & 0.91$\pm$0.0011 & 0.97$\pm$0.0001 \\
0.1 & 0.4 & 1.00$\pm$0.0002 & 1.00$\pm$0.0008 & 0.91$\pm$0.0025 & 0.97$\pm$0.0012 \\
0.1 & 0.5 & 1.00$\pm$0.0001 & 1.00$\pm$0.0005 & 0.91$\pm$0.0016 & 0.97$\pm$0.0003 \\
0.2 & 0.0 & 1.00$\pm$0.0003 & 1.00$\pm$0.0006 & 0.91$\pm$0.0018 & 0.98$\pm$0.0005 \\
0.2 & 0.1 & 1.00$\pm$0.0001 & 1.00$\pm$0.0003 & 0.91$\pm$0.0009 & 0.98$\pm$0.0004 \\
0.2 & 0.2 & 1.00$\pm$0.0004 & 1.00$\pm$0.0005 & 0.91$\pm$0.0019 & 0.98$\pm$0.0003 \\
0.2 & 0.3 & 1.00$\pm$0.0003 & 1.00$\pm$0.0007 & 0.91$\pm$0.0006 & 0.98$\pm$0.0003 \\
0.2 & 0.4 & 1.00$\pm$0.0003 & 1.00$\pm$0.0008 & 0.91$\pm$0.0014 & 0.98$\pm$0.0003 \\
0.2 & 0.5 & 1.00$\pm$0.0004 & 1.00$\pm$0.0004 & 0.91$\pm$0.0009 & 0.98$\pm$0.0006 \\
0.4 & 0.0 & 1.00$\pm$0.0001 & 1.00$\pm$0.0008 & 0.92$\pm$0.0015 & 0.99$\pm$0.0002 \\
0.4 & 0.1 & 1.00$\pm$0.0002 & 1.00$\pm$0.0009 & 0.92$\pm$0.0016 & 0.99$\pm$0.0002 \\
0.4 & 0.2 & 1.00$\pm$0.0002 & 1.00$\pm$0.0005 & 0.92$\pm$0.0013 & 0.98$\pm$0.0001 \\
0.4 & 0.3 & 1.00$\pm$0.0002 & 1.00$\pm$0.0007 & 0.91$\pm$0.0020 & 0.98$\pm$0.0003 \\
0.4 & 0.4 & 1.00$\pm$0.0001 & 1.00$\pm$0.0006 & 0.91$\pm$0.0014 & 0.98$\pm$0.0002 \\
0.4 & 0.5 & 1.00$\pm$0.0002 & 1.00$\pm$0.0002 & 0.91$\pm$0.0014 & 0.98$\pm$0.0004 \\
0.6 & 0.0 & 1.00$\pm$0.0002 & 1.00$\pm$0.0004 & 0.92$\pm$0.0014 & 0.99$\pm$0.0007 \\
0.6 & 0.1 & 1.00$\pm$0.0001 & 1.00$\pm$0.0006 & 0.92$\pm$0.0012 & 0.99$\pm$0.0005 \\
0.6 & 0.2 & 1.00$\pm$0.0004 & 1.00$\pm$0.0003 & 0.92$\pm$0.0013 & 0.99$\pm$0.0008 \\
0.6 & 0.3 & 1.00$\pm$0.0003 & 1.00$\pm$0.0004 & 0.92$\pm$0.0011 & 0.99$\pm$0.0006 \\
0.6 & 0.4 & 1.00$\pm$0.0004 & 1.00$\pm$0.0012 & 0.91$\pm$0.0013 & 0.98$\pm$0.0004 \\
0.6 & 0.5 & 1.00$\pm$0.0002 & 1.00$\pm$0.0008 & 0.91$\pm$0.0011 & 0.98$\pm$0.0007 \\
0.8 & 0.0 & 1.00$\pm$0.0001 & 1.00$\pm$0.0005 & 0.92$\pm$0.0010 & 0.99$\pm$0.0006 \\
0.8 & 0.1 & 1.00$\pm$0.0001 & 1.00$\pm$0.0006 & 0.92$\pm$0.0022 & 0.99$\pm$0.0008 \\
0.8 & 0.2 & 1.00$\pm$0.0004 & 1.00$\pm$0.0002 & 0.92$\pm$0.0009 & 0.99$\pm$0.0004 \\
0.8 & 0.3 & 1.00$\pm$0.0002 & 1.00$\pm$0.0005 & 0.92$\pm$0.0015 & 0.99$\pm$0.0002 \\
0.8 & 0.4 & 1.00$\pm$0.0002 & 1.00$\pm$0.0007 & 0.92$\pm$0.0027 & 0.99$\pm$0.0006 \\
0.8 & 0.5 & 1.00$\pm$0.0001 & 1.00$\pm$0.0007 & 0.91$\pm$0.0051 & 0.98$\pm$0.0003 \\
1.0 & 0.0 & 1.00$\pm$0.0002 & 1.00$\pm$0.0006 & 0.93$\pm$0.0007 & 0.99$\pm$0.0005 \\
1.0 & 0.1 & 1.00$\pm$0.0001 & 1.00$\pm$0.0009 & 0.92$\pm$0.0013 & 0.99$\pm$0.0004 \\
1.0 & 0.2 & 1.00$\pm$0.0004 & 1.00$\pm$0.0003 & 0.92$\pm$0.0010 & 0.99$\pm$0.0005 \\
1.0 & 0.3 & 1.00$\pm$0.0002 & 1.00$\pm$0.0005 & 0.92$\pm$0.0011 & 0.99$\pm$0.0004 \\
1.0 & 0.4 & 1.00$\pm$0.0003 & 1.00$\pm$0.0015 & 0.92$\pm$0.0017 & 0.99$\pm$0.0004 \\
1.0 & 0.5 & 1.00$\pm$0.0003 & 1.00$\pm$0.0009 & 0.91$\pm$0.0022 & 0.98$\pm$0.0002 \\
\bottomrule
\end{tabular}
}
\end{center}
\end{table}
\newpage

\begin{table}[H]
\caption{\textbf{Final Test Accuracy.} Unlike train accuracy, test accuracy showed different results depending on the subset ratio and label noise ratio, indicating the existence of loss of generalizability.} 
\vspace{2mm}
\label{table:warmstart_test}
\begin{center}
\resizebox{0.98 \textwidth}{!}{
\begin{tabular}{cccccc}
\toprule
Subset Ratio & Label Noise & MNIST (3-layer MLP) & CIFAR-10 (ResNet-18) & CIFAR-100 (ViT-Tiny) &Tiny ImageNet (VGG-16) \\
\midrule
0.1 & 0.0 & 0.98$\pm$0.0016 & 0.74$\pm$0.0023 & 0.45$\pm$0.0027 & 0.18$\pm$0.0036 \\
0.1 & 0.1 & 0.97$\pm$0.0008 & 0.72$\pm$0.0050 & 0.45$\pm$0.0032 & 0.16$\pm$0.0029 \\
0.1 & 0.2 & 0.97$\pm$0.0006 & 0.71$\pm$0.0123 & 0.44$\pm$0.0053 & 0.15$\pm$0.0039 \\
0.1 & 0.3 & 0.97$\pm$0.0012 & 0.70$\pm$0.0054 & 0.44$\pm$0.0066 & 0.14$\pm$0.0057 \\
0.1 & 0.4 & 0.97$\pm$0.0013 & 0.69$\pm$0.0095 & 0.44$\pm$0.0033 & 0.13$\pm$0.0043 \\
0.1 & 0.5 & 0.97$\pm$0.0019 & 0.69$\pm$0.0067 & 0.44$\pm$0.0089 & 0.12$\pm$0.0046 \\
0.2 & 0.0 & 0.98$\pm$0.0010 & 0.75$\pm$0.0047 & 0.45$\pm$0.0018 & 0.24$\pm$0.0008 \\
0.2 & 0.1 & 0.97$\pm$0.0009 & 0.72$\pm$0.0049 & 0.44$\pm$0.0039 & 0.22$\pm$0.0034 \\
0.2 & 0.2 & 0.97$\pm$0.0010 & 0.71$\pm$0.0047 & 0.44$\pm$0.0026 & 0.20$\pm$0.0036 \\
0.2 & 0.3 & 0.97$\pm$0.0009 & 0.69$\pm$0.0028 & 0.44$\pm$0.0032 & 0.18$\pm$0.0020 \\
0.2 & 0.4 & 0.97$\pm$0.0011 & 0.68$\pm$0.0032 & 0.44$\pm$0.0041 & 0.16$\pm$0.0039 \\
0.2 & 0.5 & 0.97$\pm$0.0012 & 0.66$\pm$0.0057 & 0.44$\pm$0.0029 & 0.15$\pm$0.0032 \\
0.4 & 0.0 & 0.98$\pm$0.0016 & 0.77$\pm$0.0038 & 0.47$\pm$0.0068 & 0.36$\pm$0.0012 \\
0.4 & 0.1 & 0.97$\pm$0.0012 & 0.75$\pm$0.0027 & 0.46$\pm$0.0047 & 0.33$\pm$0.0015 \\
0.4 & 0.2 & 0.97$\pm$0.0018 & 0.73$\pm$0.0035 & 0.44$\pm$0.0027 & 0.32$\pm$0.0026 \\
0.4 & 0.3 & 0.97$\pm$0.0013 & 0.71$\pm$0.0053 & 0.44$\pm$0.0027 & 0.30$\pm$0.0061 \\
0.4 & 0.4 & 0.97$\pm$0.0014 & 0.69$\pm$0.0066 & 0.44$\pm$0.0027 & 0.27$\pm$0.0026 \\
0.4 & 0.5 & 0.97$\pm$0.0018 & 0.67$\pm$0.0092 & 0.43$\pm$0.0043 & 0.25$\pm$0.0029 \\
0.6 & 0.0 & 0.98$\pm$0.0018 & 0.79$\pm$0.0042 & 0.48$\pm$0.0042 & 0.40$\pm$0.0032 \\
0.6 & 0.1 & 0.97$\pm$0.0009 & 0.77$\pm$0.0045 & 0.47$\pm$0.0042 & 0.38$\pm$0.0012 \\
0.6 & 0.2 & 0.97$\pm$0.0014 & 0.75$\pm$0.0024 & 0.46$\pm$0.0042 & 0.36$\pm$0.0032 \\
0.6 & 0.3 & 0.97$\pm$0.0012 & 0.73$\pm$0.0020 & 0.44$\pm$0.0065 & 0.33$\pm$0.0041 \\
0.6 & 0.4 & 0.97$\pm$0.0011 & 0.71$\pm$0.0048 & 0.44$\pm$0.0030 & 0.30$\pm$0.0016 \\
0.6 & 0.5 & 0.97$\pm$0.0010 & 0.69$\pm$0.0040 & 0.43$\pm$0.0040 & 0.27$\pm$0.0005 \\
0.8 & 0.0 & 0.98$\pm$0.0014 & 0.81$\pm$0.0045 & 0.50$\pm$0.0050 & 0.42$\pm$0.0045 \\
0.8 & 0.1 & 0.97$\pm$0.0006 & 0.78$\pm$0.0050 & 0.48$\pm$0.0076 & 0.40$\pm$0.0019 \\
0.8 & 0.2 & 0.97$\pm$0.0014 & 0.77$\pm$0.0026 & 0.46$\pm$0.0054 & 0.38$\pm$0.0024 \\
0.8 & 0.3 & 0.97$\pm$0.0014 & 0.75$\pm$0.0063 & 0.46$\pm$0.0041 & 0.35$\pm$0.0023 \\
0.8 & 0.4 & 0.97$\pm$0.0019 & 0.74$\pm$0.0033 & 0.45$\pm$0.0054 & 0.33$\pm$0.0037 \\
0.8 & 0.5 & 0.97$\pm$0.0018 & 0.73$\pm$0.0075 & 0.43$\pm$0.0093 & 0.29$\pm$0.0022 \\
1.0 & 0.0 & 0.98$\pm$0.0011 & 0.81$\pm$0.0053 & 0.51$\pm$0.0078 & 0.43$\pm$0.0009 \\
1.0 & 0.1 & 0.97$\pm$0.0012 & 0.79$\pm$0.0060 & 0.49$\pm$0.0052 & 0.41$\pm$0.0019 \\
1.0 & 0.2 & 0.97$\pm$0.0013 & 0.78$\pm$0.0071 & 0.48$\pm$0.0035 & 0.40$\pm$0.0081 \\
1.0 & 0.3 & 0.97$\pm$0.0011 & 0.77$\pm$0.0075 & 0.47$\pm$0.0052 & 0.37$\pm$0.0022 \\
1.0 & 0.4 & 0.97$\pm$0.0022 & 0.76$\pm$0.0051 & 0.45$\pm$0.0037 & 0.34$\pm$0.0029 \\
1.0 & 0.5 & 0.97$\pm$0.0010 & 0.75$\pm$0.0076 & 0.40$\pm$0.0376 & 0.30$\pm$0.0033 \\
\bottomrule
\end{tabular}
}
\end{center}
\end{table}

\newpage
\subsection{Influence of Dataset on Loss of Generalizability}
\label{appendix:dataset_architecture_impact}

\begin{figure}[h]
\begin{center}
\includegraphics[width=0.5\linewidth]{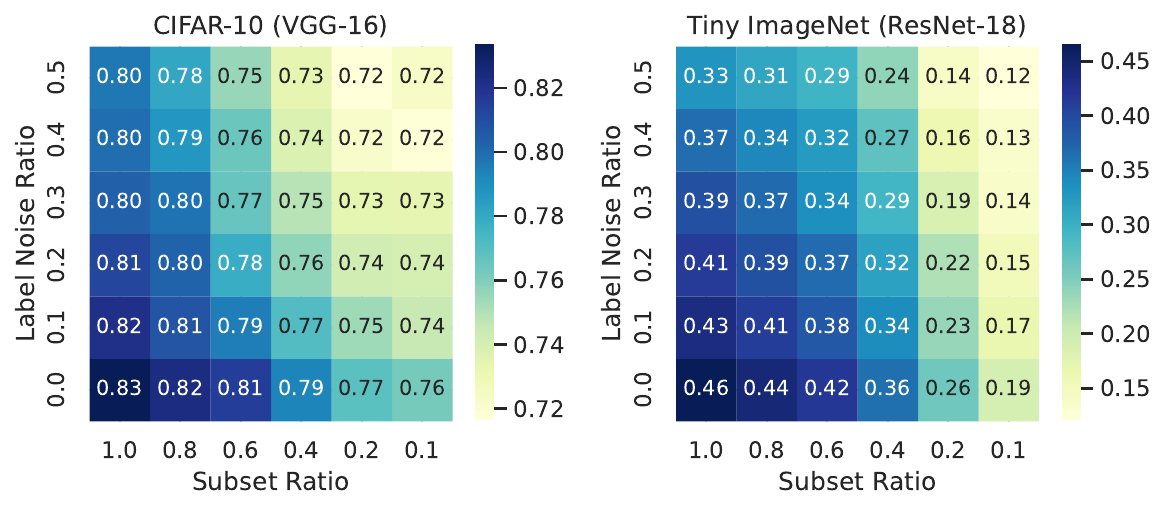}
\end{center}
\caption{\textbf{Datasets and Models Affect Plasticity.} Results when swapping models between CIFAR-10 and Tiny ImageNet in the previous setting. It highlights the similar aspects of loss of generalizability when datasets are paired with different models.}
\label{figure:data-model}
\end{figure}

In this section, as an extension of our investigation into the effects of dataset size and label noise discussed in Section \ref{subsection:size_noise_effects}, we aimed to understand the complex relationship between datasets and models in terms of generalization. To do this, we conducted experiments in which we swapped the models for the CIFAR-10 and Tiny ImageNet datasets. Specifically, we used ResNet-18 for Tiny ImageNet and VGG-16 for CIFAR-10. Each configuration was tested with two different random seeds, and the results are illustrated in Figure \ref{figure:data-model}.

Figure \ref{figure:data-model} demonstrates how the choice of datasets and models impacts plasticity. Comparing these results to the earlier findings shown in Figure \ref{figure:fig2}.(c), we observed a consistent trend across datasets. Notably, Tiny ImageNet consistently exhibited a more significant loss in generalization ability compared to CIFAR-10, regardless of the model used. This highlights the critical role of the characteristics of the dataset in understanding the loss of generalizability.

\subsection{Baseline Comparison: Train and Test Accuracies}

Below are the final train and test accuracies from our experiments in Section \ref{sub_section:generalization_effectiveness}.

\vspace{-1mm}
\begin{table*}[h]
\begin{center}
\caption{\textbf{Final Train Accuracy}.  Warm-starting has no adverse effect on train accuracy in widely used architectures, achieving near-100\% accuracy on CIFAR-10 and Tiny ImageNet. This highlights the significant gap between trainability and generalizability.}
\vspace{2mm}
\resizebox{0.98 \textwidth}{!}{
\begin{tabular}{@{\extracolsep{4pt}}l l c c c c c c c c c c c c}
\toprule

\multirow{2}{*}{Dataset} 
& \multirow{2}{*}{Architecture} 
& \multicolumn{3}{c}{Baseline} 
& \multicolumn{2}{c}{Generalizability} 
& \multicolumn{4}{c}{Trainability} 
& \multicolumn{3}{c}{Re-initialization} \\ 
\cline{3-5} \cline{6-7} \cline{8-11} \cline{12-14}\\[-2.0ex]

&
& WS   
& w/o WS  
& w/o WS + Aug  
& L2 
& Aug     
& Spectral  
& Regen 
& ReDo 
& CReLU 
& HR
& SnP
& SnP + L2 + Aug         \\
\cmidrule{1-14} \\[-2.5ex]

CIFAR-10 
& ResNet-18  
& 0.998
& 0.998
& 0.994
& 0.990
& 0.993
& 0.997
& 0.996
& 0.998
& 0.997
& 0.997
& 0.997
& 0.994
\\
CIFAR-100                
& ViT-Tiny      
& 0.907
& 0.912
& 0.852
& 0.865
& 0.867
& 0.913
& 0.906
& 0.908
& 0.890
& 0.911
& 0.915
& 0.872
\\
T-ImageNet            
& VGG-16
& 0.972
& 0.984
& 0.974
& 0.925
& 0.963
& 0.997
& 0.979
& 0.972
& 0.964
& 0.979
& 0.981
& 0.970
\\ 
\bottomrule
\label{table:warmstart_train_baseline}
\end{tabular}}
\end{center}
\end{table*}

\vspace{-6mm}

\begin{table*}[h]
\begin{center}
\caption{\textbf{Final Test Accuracy.} While trainability enhancements can yield high training accuracies, the best test performance is achieved by combining Shrink \& Perturb with generalization techniques..}
\vspace{2mm}
\resizebox{0.98 \textwidth}{!}{
\begin{tabular}{@{\extracolsep{4pt}}l l c c c c c c c c c c c c}
\toprule

\multirow{2}{*}{Dataset} 
& \multirow{2}{*}{Architecture} 
& \multicolumn{3}{c}{Baseline} 
& \multicolumn{2}{c}{Generalizability} 
& \multicolumn{4}{c}{Trainability} 
& \multicolumn{3}{c}{Re-initialization} \\ 
\cline{3-5} \cline{6-7} \cline{8-11} \cline{12-14}\\[-2.0ex]

&
& WS   
& w/o WS  
& w/o WS + Aug   
& L2 
& Aug     
& Spectral  
& Regen 
& ReDo 
& CReLU 
& HR
& SnP
& SnP + L2 + Aug         \\
\cmidrule{1-14} \\[-2.5ex]

CIFAR-10 
& ResNet-18  
& 0.704
& 0.835
& 0.924
& 0.735
& 0.840
& 0.705
& 0.701
& 0.682
& 0.678
& 0.693
& 0.771
& \textbf{0.910}
\\
CIFAR-100                
& ViT-Tiny      
& 0.442
& 0.500
& 0.555
& 0.457
& 0.511
& 0.444
& 0.447
& 0.445
& 0.440
& 0.445
& 0.467
& \textbf{0.536}
\\
T-ImageNet            
& VGG-16& 0.115
& 0.439
& 0.463
& 0.221
& 0.227
& 0.118
& 0.127
& 0.118
& 0.118
& 0.177
& 0.229
& \textbf{0.376}
\\ 
\bottomrule
\label{table:warmstart_test_baseline}
\end{tabular}}
\end{center}
\end{table*}

In all experiments, training accuracy was on par with a freshly initialized model (w/o WS) for most methods, except for those aimed at preventing overfitting.
Table \ref{table:warmstart_test} displays the final test accuracy results. In these experiments, trainability methods had limited impact, despite the increase in training accuracy in the Spectral Decomposition method.
While generalizability methods (L2, Data Augmentation) generally improved performance compared to the baseline (warm-start), the performance was still significantly lower than freshly initialized models, indicating that these methods couldn't fully mitigate plasticity loss. Shrink \& Perturb with generalizability methods outperformed all the other approaches, with its performance reaching close to training the fresh networks.

\section{Relationship between Generalizability and Potential Indicators} 
\label{appendix:metrics}

The precise mechanisms underlying the loss of plasticity in neural networks remain incompletely understood. However, several factors have been proposed as potential indicators of this phenomenon. These factors include weight magnitude, the distance of current weight values from their initial values, the number of zero gradients or non-active ReLU units, and feature rank collapse, as highlighted in prior studies \citep{dohare2021continual, zilly2021plasticity, dohare2023maintaining, kumar2023continual, kumar2023regen, abbas2023plasticity_crelu, lyle2023understanding_plasticity, lewandowski2023curvature}.

\begin{figure}[h]
\begin{center}
\includegraphics[width=0.9\linewidth]{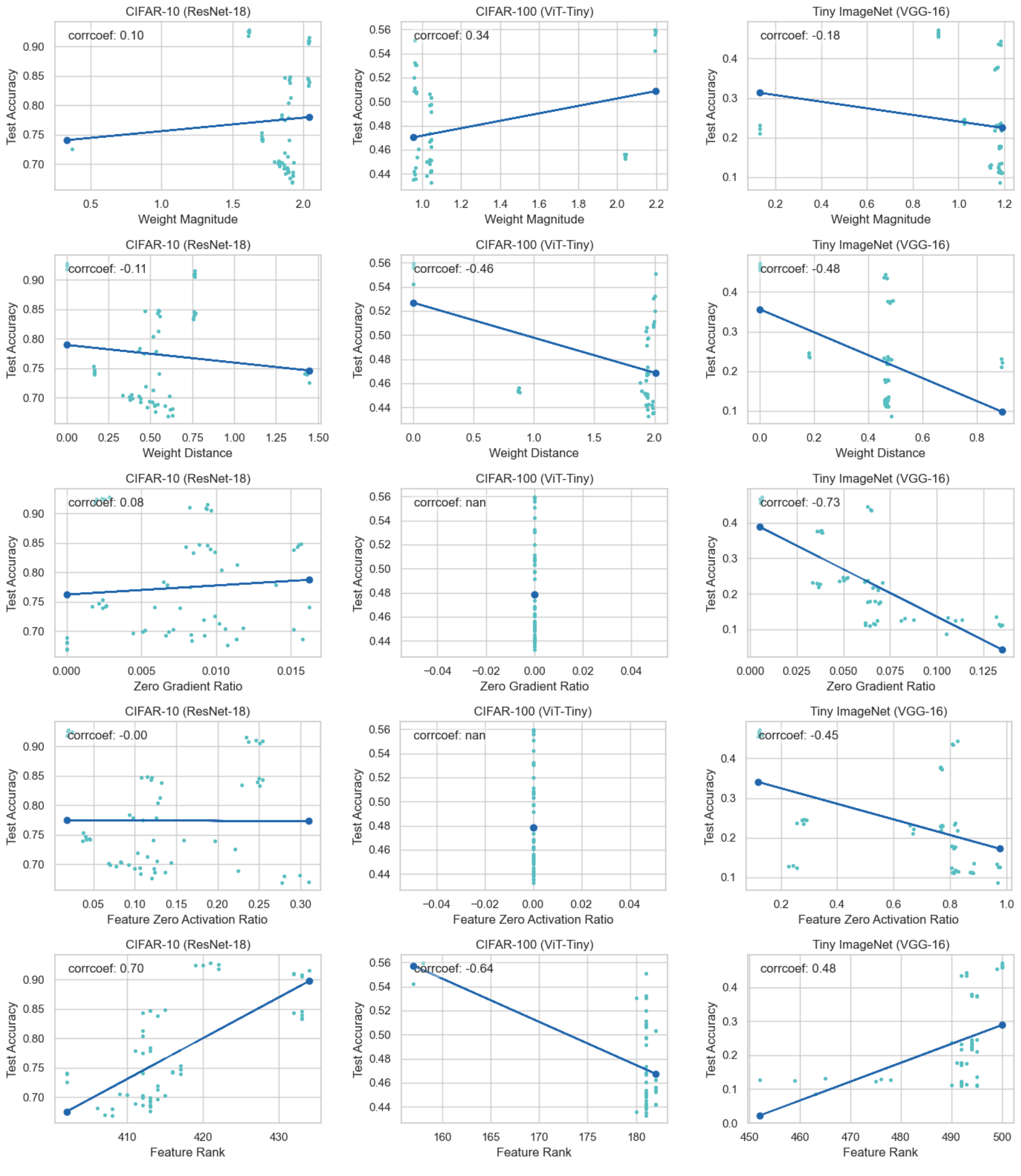} 
\end{center}
\caption{\textbf{Metrics}. None of the metrics exhibit a strong and consistent correlation to generalization performance.}

\label{figure:metric}
\end{figure}

This study delves into the relationship between these metrics and the generalization ability of neural networks. Specifically, we focus on examining the Pearson correlation coefficients between these metrics and the final test accuracy, with each metric being measured at the end of the warm-starting phase. As depicted in Figure \ref{figure:metric}, we observed that none of these metrics exhibit a strong and consistent correlation across different datasets and architectures, although weight distance showed a subtle trend. Below, we provide details on how each metric is calculated:

\textbf{Weight Magnitude:} This metric is calculated as the cumulative average of the squared weights across all layers of the model, given by the formula:
\[
\text{norm}(\mathbf{W}) = \sum_{l \in L} \frac{1}{\text{count}(l)} \sum_{w \in l} w^2
\]
Here, \( L \) stands for the collection of layers, and \( w \) denotes the individual weights within layer \( l \).

\textbf{Weight Distance:} It measures the mean squared distance between the current weights and their initial values for each layer:
\[
\text{dist}(\mathbf{W}) = \sum_{l \in L} \frac{1}{\text{count}(l)} \sum_{w \in l} (w - w_0)^2
\]
where \( w_0 \) represents the initial weight value.

\textbf{Zero Gradient Ratio:} Indicates the percentage of zero gradients in the features produced by the backbone network's output.

\textbf{Feature Zero Activation Ratio:} Measures the proportion of zero activations in the feature output of the backbone network.

\textbf{Feature Rank (srank):}  Introduced by \citet{kumar2020implicit}, this metric assesses the effective number of unique bases in the feature matrix (\(\Phi\)), which we calculated from the output of the backbone network. Given the ordered set of singular values \(\sigma_{1}(\mathbf{\Phi}) > \sigma_{2}(\mathbf{\Phi}) > ... > \sigma_{n}(\mathbf{\Phi})\), we measure it as:
\[
    \text{srank}(\mathbf{\Phi}) = \min_k \left( \frac{\sum_{i=1}^k \sigma_i(\mathbf{\Phi})}{\sum_{j=1}^n \sigma_j(\mathbf{\Phi})} \right) \geq 1 - \delta
\]
where \(\delta\) is set to 0.01, following from \citet{kumar2020implicit}.

\clearpage

\section{Ablation Studies}
\label{appendix:ablation_studies}
In our ablation studies, we explore the robustness of our proposed method by varying two hyperparameters: momentum value and reset interval.

\begin{figure}[h]
\begin{center}
\includegraphics[width=0.99\linewidth]{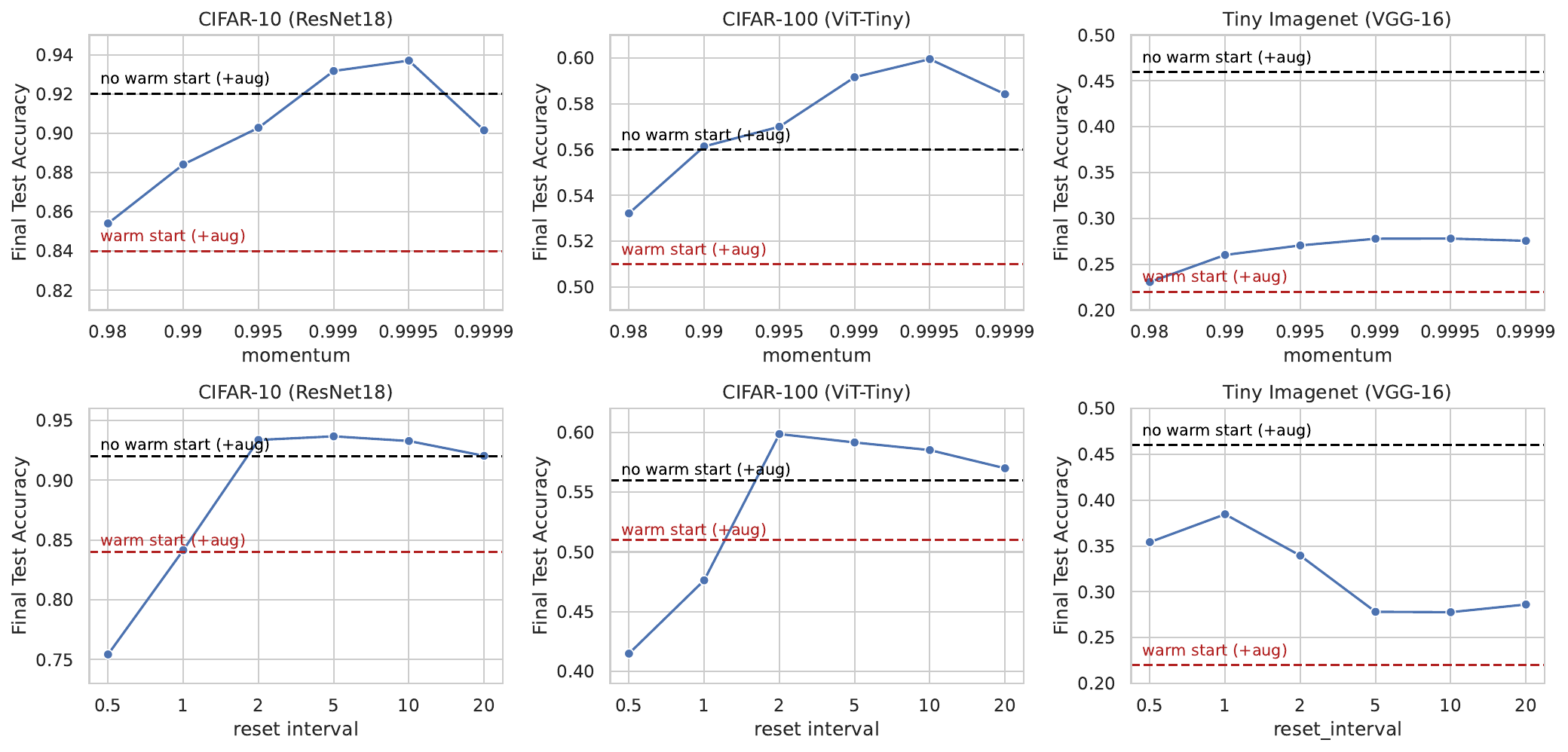}
\end{center}
\vspace{-2mm}
\caption{\textbf{Ablation Study.} The figure illustrates the robust performance of our method under variations in (a) momentum value and (b) reset interval.}
\label{figure:ablation}
\end{figure}

Our main observation is that regardless of the specific values chosen for these hyperparameters, our method consistently delivers stable and improved performance compared to the warm-starting model. This stability and performance enhancement are consistent across various datasets and architectures.

\clearpage

\section{Reinforcement Learning Results}

\begin{table}[ht]
\begin{center}
\caption{\textbf{Mean trajectory scores on Atari-100k Benchmark.} We report the individual scores on the 26 Atari games. 
All the other experiments, including DrQ, were conducted based on their original code and averaged over 5 random seeds with a replay ratio of 2.}
\vspace{2.0ex}
\resizebox{0.8\textwidth}{!}{
\begin{tabular}{l|cccccc|ccc}
\toprule &\\[-2.5ex]
Method      & \multicolumn{6}{c|}{DrQ}  & \multicolumn{3}{c}{BBF} \\[0.2ex]
\midrule \\[-2.7ex]
S\&P           & -             & \checkmark & -         & -            & -          & -          & \checkmark & \checkmark          & \checkmark \\[0.2ex]
HR             & -             & -          & -         & 20k          & 40k        & 40k        & \checkmark & \checkmark          & \checkmark \\[0.2ex]
H\&T           & -             & -          &\checkmark & -            & -          & \checkmark & -          & \checkmark          & - \\[0.2ex]
SSL            & -             & -          & -         & -            & -          & -          & -          & -                   & \checkmark \\[0.2ex]
\midrule \\[-2.7ex]
Alien          & $841.52$      & $729.1$ & $730.06$ & $838.6$ & $947.22$ & $924.9$               & $1228.7$   & $1180.0$            & $1121.7$  \\
Amidar         & $182.46$      & $122.78$ & $148.82$ & $181.26$ & $166.01$ & $221.67$            & $219.8$    & $ 293.1$             & $236.6$   \\
Assault        & $559.98$      & $848.25$ & $604.08$ & $765.18$ & $653.28$ & $638.26$            & $1657.3$   & $1844.2$             & $2004.5$  \\
Asterix        & $831.4$       & $572.5$ & $823.5$ & $791.8$ & $813.4$ & $819.3$                 & $4699.0$   & $4121.6$            & $3169.7$  \\
BankHeist      & $138.94$      & $24.74$ & $263.56$ & $51.56$ & $86.78$ & $56.32$                 & $315.4$    & $712.8$             & $768.8$   \\
BattleZone     & $6404.0$      & $2762.0$ & $8038.0$ & $4040.0$ & $6186.0$ & $4694.0$            & $23752.1$  & $ 21638.5$           & $ 23681.4$ \\
Boxing         & $10.6$        & $42.84$ & $14.29$ & $40.98$ & $23.61$ & $34.26$                & $60.8$     & $62.5$              & $77.3$    \\
Breakout       & $11.95$       & $9.8$ & $13.13$ & $18.91$ & $12.55$ & $15.69$                   & $245.6$    & $223.3$              & $331.0$    \\
CrazyClimber   & $12031.2$     & $6364.0$ & $13808.4$ & $16158.0$ & $14649.4$ & $18154.2$        & $56440.7$   & $45579.0$             & $60864.5$   \\
ChopperCommand & $917.2$       & $600.6$ & $673.4$ & $976.0$ & $922.6$ & $816.6$                & $2149.3$  & $3238.0$           & $4251.5$ \\
DemonAttack    & $663.93$      & $906.47$ & $657.41$ & $1013.14$ & $819.02$ & $813.4$            & $11502.5$  & $8248.7$            & $ 18298.3$  \\
Freeway        & $27.28$       & $21.99$ & $27.29$ & $28.33$ & $29.21$ & $28.95$                & $21.4$     & $21.9$              & $23.1$    \\
Frostbite      & $828.96$      & $561.16$ & $2218.9$ & $1262.06$ & $1923.42$ & $1939.32$        & $ 2266.2$   & $2838.2$            & $2023.0$  \\
Gopher         & $397.52$      & $602.24$ & $521.76$ & $533.32$ & $577.84$ & $455.76$            & $1758.3$    & $2068.1$             & $1209.4$   \\
Hero           & $7198.58$     & $6293.84$ & $6193.58$ & $7119.41$ & $7677.46$ & $6367.63$      & $ 4905.8$  & $7448.5$           & $ 5741.8$  \\
Jamesbond      & $349.0$       & $192.5$ & $317.5$ & $442.6$ & $281.6$ & $358.2$                & $753.1$    & $921.7$             & $1124.6$   \\
Kangaroo       & $3582.0$      & $614.4$ & $2685.2$ & $5534.8$ & $2922.8$ & $2211.4$            & $5584.2$  & $6366.7$            & $5032.0$  \\
Krull          & $3996.46$     & $4697.5$ & $3577.46$ & $4414.6$ & $3897.84$ & $3920.04$         & $ 7504.2$   & $7695.1$            & $ 8069.8$  \\
KungFuMaster   & $17319.6$     & $5106.2$ & $12035.4$ & $13909.2$ & $13032.4$ & $19138.8$       & $18166.8$  & $17296.7$           & $16616.8$ \\
MsPacman       & $1251.24$     & $849.66$ & $1424.3$ & $1106.18$ & $1459.34$ & $1062.72$        & $1941.0$  & $1749.3$            & $2217.8$  \\
Pong           & $-7.13$       & $-20.34$ & $-15.01$ & $-10.82$ & $-13.96$ & $-4.11$            & $13.0$     & $12.8$               & $13.6$    \\
Qbert          & $2304.05$     & $763.05$ & $3382.6$ & $3188.45$ & $3620.3$ & $3408.35$         & $3530.5 $  & $4237.0$            & $3245.3$  \\
RoadRunner     & $12821.6$     & $6594.8$ & $14674.2$ & $18029.0$ & $16383.0$ & $13995.6$        & $27572.1$  & $29422.4$           & $26419.0$ \\
Seaquest       & $498.12$      & $390.88$ & $531.92$ & $564.08$ & $457.6$ & $550.56$            & $1188.1$    & $1168.8$             & $988.6$   \\
PrivateEye     & $80.0$        & $33.0$ & $85.84$ & $34.62$ & $82.4$ & $100.0$                  & $34.8$     & $28.5$             & $39.0$   \\
UpNDown        & $4214.26$     & $4420.98$ & $4269.64$ & $9608.54$ & $6508.24$ & $8451.4$        & $10266.5$ & $7742.0$           & $15122.6$ \\
\midrule \\[-2.7ex]
IQM            & $0.243$         & $0.139$   & $0.287$    & $0.332$      & $0.288$    & $0.328$    & $0.826$    & $0.891$             & $0.940$   \\
Median         & $ 0.193$       & $0.138$   & $0.260$    & $0.254$        & $0.241$      & $0.329$    & $0.711$    & $0.749$             & $0.755$   \\
Mean           & $0.468$       & $0.458$   & $0.471$    & $0.694$        & $0.532$      & $0.584$    & $1.737$    & $1.719$             & $2.175$   \\
OG             & $0.642$         & $0.728$   & $0.617$    & $0.580$      & $0.607$    & $0.583$    & $0.397$    & $0.372$             & $0.377$   \\
\bottomrule 
\end{tabular}}
\end{center}
\label{table:atari_full_result}
\end{table}


\end{document}